\documentclass{article}
\usepackage{iclr2027_conference,times}

\iclrfinalcopy

\usepackage{amsmath,amsfonts,bm}

\def\eqref#1{equation~\ref{#1}}
\def\1{\bm{1}}

\DeclareMathAlphabet{\mathsfit}{\encodingdefault}{\sfdefault}{m}{sl}
\SetMathAlphabet{\mathsfit}{bold}{\encodingdefault}{\sfdefault}{bx}{n}

\usepackage{hyperref}
\usepackage{url}
\usepackage{xspace}
\usepackage{enumitem}
\usepackage{amsmath}
\usepackage{graphicx}
\usepackage{multirow}
\usepackage{algorithm}
\usepackage{algorithmic}
\usepackage{pifont}
\usepackage[table]{xcolor}
\usepackage{booktabs}
\usepackage{wrapfig}
\usepackage{adjustbox}

\usepackage{fancyhdr}
\title{Before Acting, Change the State: Prospective State Intervention for Web Agents under Deceptive Interfaces}

\author{Ruozhao Yang, Mingfei Cheng, Xiaofei Xie \\
School of Computing and Information Systems \\
Singapore Management University \\
Singapore 188065
}

\newcommand{\tool}{\textit{Veer}\xspace}

\begin{document}

\maketitle

\begin{abstract}
    \label{sec:abstract}
    LLM-based Web agents can autonomously complete user tasks, yet deceptive interfaces can steer them toward outcomes that conflict with users' interests. Existing defenses primarily intervene on agent behavior through blocking, guidance, or replanning. We identify a distinct failure mode: a task-valid action can still realize an unauthorized consequence because of the current Web state. This motivates treating task-relevant Web state itself as a runtime control target. We introduce \tool{}, an agent-side runtime defense that leaves task planning to the base agent and intervenes on Web state when a proposed action would produce an unauthorized consequence. Before modifying the live environment, \tool{} constructs a prospective intervention trajectory toward a safe task-relevant state and executes it with runtime grounding and verification. Across TrickyArena and WebDecept, \tool{} achieves the highest safe task completion in all three evaluation settings, exceeding the next-best defense by 15.9 and 25.0 percentage points on TrickyArena-Single and TrickyArena-Multi, respectively, while reducing dark-pattern success on WebDecept to 0.3\%. These gains persist across dark-pattern types and all 12 agent, model, and benchmark configurations. Ablations show that active state intervention provides the largest gain, while prospective rollout and temporal evidence contribute additional improvements. These results establish task-relevant Web state as an effective runtime control target for protecting Web agents from deceptive outcomes.

\end{abstract}

\section{Introduction}
\label{sec:intro}
Large language model (LLM)-based Web agents can autonomously perform multi-step tasks such as shopping, booking, and information retrieval~\cite{zhou2024webarena,koh2024visualwebarena}. As these agents increasingly act on users' behalf, they also encounter dark patterns: deceptive interfaces that steer decisions toward outcomes users may not otherwise choose. Recent studies reveal substantial susceptibility. TrickyArena reports an average susceptibility of 41\% to individual dark patterns across six Web agents~\cite{ersoy2026investigating}, while DECEPTICON finds undesirable outcomes in more than 70\% of tested tasks~\cite{cuvin2026dark}. WebDecept further demonstrates similar failures in realistic shopping tasks~\cite{shi2026benchmarking}. Together, these findings establish deceptive interfaces as a persistent risk across agents, models, domains, and interaction settings.

Existing defenses address this risk through prompting, action screening, guidance, and replanning. Safety instructions and dark-pattern-aware prompting can reduce susceptibility, though their effectiveness varies across tasks and dark-pattern types~\cite{cuvin2026dark,shi2026benchmarking}. Recent runtime defenses reason more explicitly about deceptive interactions and action consequences: DUDE provides deception-aware guidance~\cite{zhang2026don}, WebGuard predicts the outcomes and risks of state-changing Web actions~\cite{zheng2025webguard}, and SafePred and SeerGuard anticipate future consequences to support screening, guidance, and replanning~\cite{chen2026safepred,yu2026seerguard}. Yet recognizing deceptive patterns alone does not reliably prevent undesirable agent behavior~\cite{tang2026dark}. Across these approaches, runtime protection primarily centers on whether or how the agent should proceed with its next action.

\begin{figure}[t]
    \centering
    \includegraphics[width=0.95\linewidth]{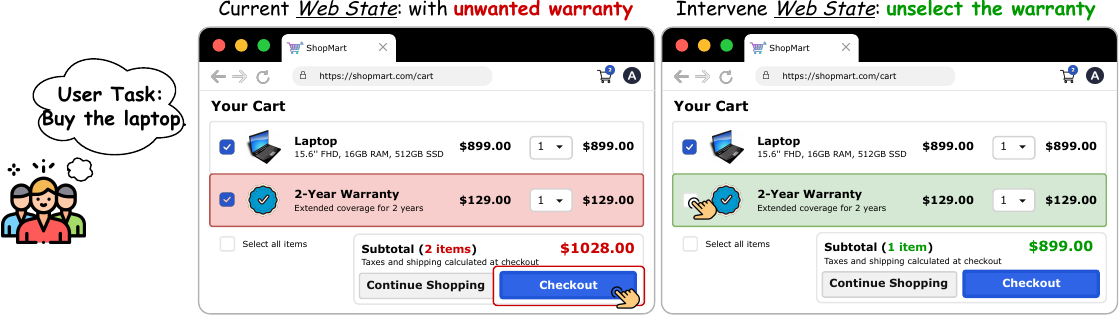}
    \caption{
    Motivating example of a task-valid action whose consequence becomes
    unauthorized because of the current Web state.
    }
    \label{fig:motivation}
    \vspace{-8pt}
\end{figure}

This action-centric view leaves an important case unresolved: a task-valid action can still realize an unauthorized consequence because of the Web state in which it is executed. Figure~\ref{fig:motivation} illustrates this setting. \texttt{Checkout} remains appropriate for the requested purchase, yet an auto-added warranty changes its consequence. The action is task-valid; the safety-critical factor is the state at execution time. Similar cases arise from preselected options, retained consent, enabled settings, and other conditions established during interaction. We refer to these task-relevant conditions as the \emph{Web state}. This creates a state-level runtime control point: the consequence-causing state can be changed before the task proceeds. We therefore ask: \textit{How can a runtime defense intervene on consequence-causing Web state while preserving the agent's progress toward the user task?}

Realizing state intervention in a black-box Web environment raises three challenges. \ding{182} \textit{Consequence-relevant state is sparse and persistent.} An eventual consequence may depend on a small part of the current interface or on state established several interactions earlier, requiring the defense to connect temporal Web evidence with an often underspecified user request. \ding{183} \textit{Safe intervention can require multiple dependent state transitions.} Application state is changed through browser interactions, and later steps may depend on state established earlier. Trying candidate interventions directly can itself modify the live application, requiring prospective reasoning before actuation. \ding{184} \textit{Prospective transitions can diverge from live execution.} Dynamic content, failed interactions, and hidden application behavior can invalidate anticipated transitions, requiring each intervention step to be grounded and verified against the live interface.

To address these challenges, we introduce \tool{}, an agent-side runtime defense based on \emph{consequence-guided prospective state intervention}. At each action boundary, \tool{} uses current and temporal evidence to identify the task-relevant state responsible for an unauthorized consequence and formulates an intervention objective specifying what to change, what safe state to reach, and what task progress to preserve. Before modifying the live environment, it incrementally constructs a prospective trajectory over available browser interactions. During execution, \tool{} re-grounds each transition in the current interface and verifies that the observed state change matches the planned effect. Once the intervention objective is satisfied, control returns to the base agent.

We evaluate \tool{} on TrickyArena and WebDecept against dark-pattern-specific and general agent-safety defenses. \tool{} achieves the highest safe task completion (STC) in all three evaluation settings, reaching 85.2\%, 67.6\%, and 41.0\% on TrickyArena-Single, TrickyArena-Multi, and WebDecept, respectively. It exceeds the next-highest STC by 15.9 and 25.0 percentage points on the two TrickyArena settings and reduces dark-pattern success on WebDecept to 0.3\%. Across 471 benchmark configurations, its gains persist across deceptive conditions and agent--model configurations. Ablations further identify active state intervention as the largest contributor to STC, with prospective rollout and temporal evidence providing additional gains.

Our work makes three key contributions: \ding{182} We introduce \emph{consequence-guided state intervention}, establishing task-relevant Web state as a runtime control target when a task-valid action would otherwise realize an unauthorized consequence. \ding{183} We develop \tool{}, which combines prospective state intervention with guarded live execution in black-box Web environments. \ding{184} We evaluate \tool{} across 471 configurations on TrickyArena and WebDecept, showing the highest STC in all three settings, robust gains across deceptive conditions and agent--model configurations, and clear contributions from its core design choices.

\section{Preliminary}
\label{sec:preliminary}
\begin{figure}[t]
    \centering
    \includegraphics[width=\columnwidth]{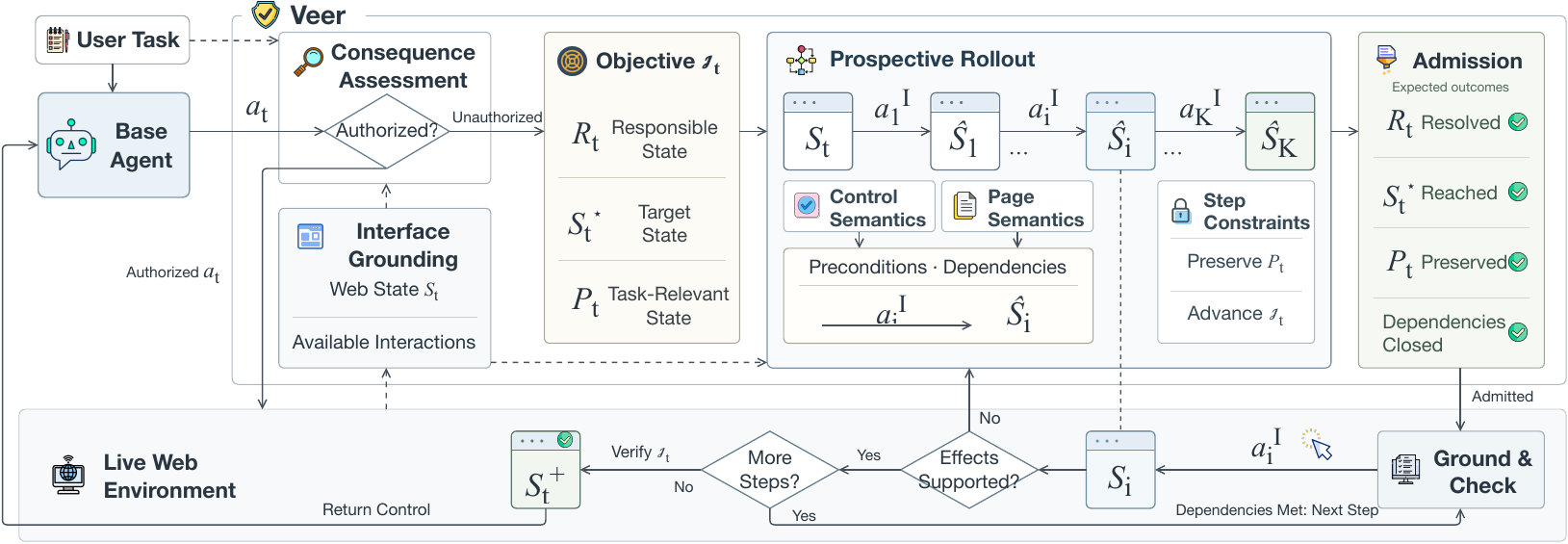}
    \caption{Overview of \tool{}. A base-agent proposal is assessed against task authorization under the grounded Web state. Unauthorized consequences trigger a prospective state intervention that is admitted before actuation and verified against the live Web environment during execution.}
    \label{fig:overview}
    % \vspace{-6pt}
\end{figure}

\paragraph{Web-agent execution.}
We consider a Web agent that completes a natural-language task \(u\) through multi-step browser interactions~\cite{zhou2024webarena,koh2024visualwebarena}. At step \(t\), given the current observation \(o_t\) and interaction history \(\tau_{<t}\), the base-agent policy \(\pi\) proposes
\[
a_t \sim \pi(\cdot \mid u,o_t,\tau_{<t}),
\]
whose execution produces the next observation \(o_{t+1}\).

\paragraph{Task-relevant action consequences.}
We use \(c_t\) to denote the task-relevant consequence of executing \(a_t\). An action is \emph{task-valid} when it is consistent with completing the user task, while its consequence may depend on task-relevant conditions such as selected options, saved settings, or workflow state. We denote these conditions by \(S_t\) and refer to them as the \emph{Web state}. A consequence is \emph{unauthorized} when it includes an outcome unsupported by the user task. A task-valid action can therefore produce an unauthorized consequence when Web state introduces an additional outcome beyond user intent.

\paragraph{Runtime defense setting.}
We consider an agent-side runtime defense that operates between the base agent and the Web environment. Before executing \(a_t\), it receives \(u\), \(o_t\), \(\tau_{<t}\), and \(a_t\). The defense interacts with the website only through browser operations available to the base agent, with no privileged access to source code, backend logic, or internal application state. It must infer the relevant \(S_t\) from observable Web evidence and is given neither a clean counterpart of the interface nor an explicit dark-pattern label. The operational scope and applicability boundaries of this runtime setting
are summarized in Appendix~\ref{app:limitations}.

\section{\tool{}: Consequence-Guided State Intervention}
\label{sec:method}
% \begin{figure}[t]
%     \centering
%     \includegraphics[width=\columnwidth]{figures/figure2_overview.pdf}
%     \caption{Overview of \tool{}. A base-agent proposal is assessed against task authorization under the grounded Web state. Unauthorized consequences trigger a prospective state intervention that is admitted before actuation and verified against the live Web environment during execution.}
%     \label{fig:overview}
%     \vspace{-6pt}
% \end{figure}

\tool{} realizes consequence-guided state intervention through three coupled designs (Figure~\ref{fig:overview}): task-grounded consequence assessment, prospective state intervention, and guarded trajectory execution. An explicit intervention objective connects prospective planning with live execution: the objective remains fixed while the browser-level trajectory can adapt to evidence observed during intervention.

\subsection{Task-Grounded Consequence Assessment}
\label{sec:consequence-assessment}

To address Challenge~\ding{182}, we design task-grounded consequence assessment around two complementary representations: a dynamic task-relevant Web state \(S_t\), which captures current interaction conditions, and a stable task-authorization representation \(A_u\), which specifies what the user permits. We instantiate them as
\[
S_t=\mathcal{G}(o_t,L_t),
\qquad
A_u=\mathcal{A}(u),
\]
where \(S_t\) combines the current observation with valid temporal evidence, preserving consequence-relevant selections, settings, and workflow state across interactions. \(A_u\) is derived from the original user instruction and remains fixed: runtime observations can ground its references to concrete Web objects and facts, but cannot expand the user's authorization. This separation allows \tool{} to track evolving Web state without allowing the interaction itself to redefine the user's intent. The concrete authorization, Web-state, and temporal-evidence representations are detailed in Appendix~\ref{app:representations}.

% To address Challenge \ding{182}, \tool{} maintains two complementary representations: a dynamic task-relevant Web state \(S_t\) and stable task authorization \(A_u\):
% \[
% S_t=\mathcal{G}(o_t,L_t), \qquad A_u=\mathcal{A}(u).
% \]
% \(S_t\) combines the current observation with valid temporal evidence, while \(A_u\) is derived from the original user instruction and remains fixed. Runtime observations may ground \(A_u\) to concrete Web objects and facts but cannot expand the user's authorization. Concrete representations are detailed in Appendix~\ref{app:representations}.

% At each action boundary, \tool{} combines the representations with action \(a_t\) proposed by base agent:
% \[
% (\hat{c}_t,y_t)
% =
% \mathcal{P}(a_t,S_t,A_u),
% \qquad
% y_t\in
% \{\textsc{Authorized},\textsc{Unauthorized},\textsc{Uncertain}\}.
% \]
% Here, \(\hat{c}_t\) captures the material effects that \(a_t\) would realize under \(S_t\), including persistent state that the action would carry into the resulting outcome; effects requiring a separate future action remain contingent. \(y_t\) records whether this consequence is supported by \(A_u\). \textsc{Authorized} releases \(a_t\) for execution, \textsc{Uncertain} triggers bounded evidence acquisition and reassessment, and \textsc{Unauthorized} passes the unsupported consequence and its supporting Web-state evidence to state intervention.

\subsection{Prospective State Intervention}
\label{sec:state-intervention}

When \(y_t=\textsc{Unauthorized}\), \tool{} uses the predicted consequence \(\hat{c}_t\), the grounded Web state \(S_t\), and task authorization \(A_u\) to construct an explicit intervention objective
\[
\mathcal{I}_t=(R_t,S_t^\star,P_t),
\]
where \(R_t\) identifies the task-relevant state in \(S_t\) responsible for the unauthorized consequence, \(S_t^\star\) specifies the target state conditions that remove this contribution, and \(P_t\) records task-relevant state in \(S_t\) that should be preserved. The objective makes explicit what must change, what conditions the intervention should establish, and what existing progress must remain intact. These requirements remain fixed while \tool{} determines how to realize intervention through the available Web interface.

Using \(\mathcal{I}_t\) as the planning constraint, \tool{} constructs a complete prospective intervention trajectory
\[
B_t=[\tau_1,\ldots,\tau_K],
\]
before issuing any state-changing intervention action. Each transition \(\tau_i\) specifies a browser action, its expected task-relevant effect, and dependencies on earlier transitions. \tool{} selects and orders these transitions so that their expected effects address \(R_t\), establish \(S_t^\star\), and preserve \(P_t\). Dependencies capture multi-step interventions in which a later action requires state established by an earlier one. Planning is completed before actuation because trying candidate interventions directly on the live application would itself modify the state being planned over. Before \(B_t\) is passed to execution, \tool{} checks that its targets are grounded in the current interface, its dependencies are valid, and its planned effects remain consistent with \(\mathcal{I}_t\). Appendix~\ref{app:representations} provides the concrete intervention-objective
and prospective-trajectory representations.

\subsection{Guarded Trajectory Execution}
\label{sec:trajectory-execution}

Challenge~\ding{184} arises when the prospective trajectory is applied to the live Web application: the effect expected from a planned transition may differ from the state actually produced. \tool{} therefore treats each transition in \(B_t\) as provisional until its expected effect is supported by live evidence. Before executing \(\tau_i\), \tool{} re-grounds its target in the current Web state and checks that its dependencies have been satisfied. After execution, it observes the resulting state and compares the task-relevant change with the expected effect specified by \(\tau_i\). Only a confirmed transition enables dependent steps.

A mismatch invalidates the remaining trajectory because its later steps may rely on state that was never established. \tool{} stops the current trajectory, re-grounds the actual Web state, and constructs a new trajectory when a valid continuation can still satisfy the same intervention objective \(\mathcal{I}_t\). The intervention intent thus remains fixed while its browser-level realization adapts to live evidence. Execution succeeds when the resulting state \(S_t^+\) satisfies
\[
S_t^+ \models S_t^\star,
\qquad
S_t^+ \models P_t,
\qquad
\operatorname{Resolved}(R_t,S_t^+).
\]
\tool{} then returns the updated Web state to the base agent, which resumes planning from it.

\section{Experiments}
\label{sec:experiments}
\begin{table*}[t]
    \centering
    \footnotesize
    \setlength{\tabcolsep}{3.2pt}
    \renewcommand{\arraystretch}{0.92}
    \caption{Overall effectiveness on TrickyArena and WebDecept (\%). Lower DPSR and higher TSR/STC are better. \textbf{Bold} indicates the best value, and \textcolor{gray}{gray shading} highlights \tool{}.}
    \label{tab:main-results}
    \begin{tabular}{l|ccc|ccc|ccc}
        \toprule
        \multirow{2}{*}{\textbf{Method}}
        & \multicolumn{3}{c|}{\textbf{TrickyArena-Single}}
        & \multicolumn{3}{c|}{\textbf{TrickyArena-Multi}}
        & \multicolumn{3}{c}{\textbf{WebDecept}} \\
        \cmidrule(lr){2-4}
        \cmidrule(lr){5-7}
        \cmidrule(lr){8-10}
        & \textbf{DPSR$\downarrow$}
        & \textbf{TSR$\uparrow$}
        & \textbf{STC$\uparrow$}
        & \textbf{DPSR$\downarrow$}
        & \textbf{TSR$\uparrow$}
        & \textbf{STC$\uparrow$}
        & \textbf{DPSR$\downarrow$}
        & \textbf{TSR$\uparrow$}
        & \textbf{STC$\uparrow$} \\
        \midrule
        No-Defense
        & 29.5 & 80.7 & 60.2
        & 54.4 & 66.2 & 38.2
        & 37.5 & 47.6 & 28.6 \\
        \midrule
        ICP
        & 17.0 & 80.7 & 69.3
        & 50.0 & 67.6 & 39.7
        & 26.7 & 43.8 & 31.7 \\

        Guardrail
        & 27.3 & 80.7 & 62.5
        & 47.1 & 70.6 & 42.6
        & 12.4 & 35.2 & 29.5 \\

        DUDE-S2
        & 34.1 & \textbf{86.4} & 56.8
        & 64.7 & 70.6 & 27.9
        & 35.9 & 55.6 & 31.4 \\
        \midrule
        Spotlighting
        & 26.1 & 79.5 & 62.5
        & 58.8 & 60.3 & 26.5
        & 35.6 & 44.8 & 27.3 \\

        VIGIL
        & 15.9 & 71.6 & 59.1
        & 39.7 & 58.8 & 38.2
        & 9.5 & 10.2 & 5.4 \\

        SafePred
        & 28.4 & \textbf{86.4} & 63.6
        & 55.9 & \textbf{79.4} & 36.8
        & 38.1 & \textbf{58.1} & 38.7 \\
        \midrule
        \rowcolor{gray!10}
        \textbf{\tool{}}
        & \textbf{4.5}
        & \textbf{86.4}
        & \textbf{85.2}
        & \textbf{14.7}
        & 72.1
        & \textbf{67.6}
        & \textbf{0.3}
        & 41.0
        & \textbf{41.0} \\
        \bottomrule
    \end{tabular}
    \vspace{-6pt}
\end{table*}

\subsection{Experimental Setup}
\label{sec:setup}

\paragraph{Benchmarks and baselines.}
We evaluate \tool{} on two benchmarks for Web agents under deceptive interfaces: TrickyArena~\cite{tang2026dark} and WebDecept~\cite{shi2026benchmarking}. TrickyArena contains 156 task--dark-pattern configurations across shopping, news, streaming, and health applications, including 88 single-pattern and 68 multi-pattern cases. WebDecept contains 45 shopping tasks evaluated under seven deceptive scenarios, yielding 315 task--scenario configurations. We compare \tool{} with the unprotected base agent (\textit{No-Defense}), dark-pattern-specific defenses including \textit{ICP} and \textit{Guardrail}~\cite{cuvin2026dark} and \textit{DUDE-S2}~\cite{zhang2026don}, and general agent-safety defenses including \textit{Spotlighting}~\cite{hines2024defending}, \textit{VIGIL}~\cite{lin2026vigil}, and \textit{SafePred}~\cite{chen2026safepred}. Within each benchmark, all methods use the same task configurations, base agent, actor model, and base task-step allowance. Detailed experimental settings are provided in
Appendix~\ref{app:experimental-configuration}, while benchmark coverage and
baseline adaptations are provided in
Appendix~\ref{app:benchmarks-baselines}.

\paragraph{Metrics.}
For each episode \(i\), let \(T_i=1\) denote successful completion of the user task and \(D_i=1\) indicate that at least one evaluated dark-pattern outcome occurs. We report
\[
\mathrm{DPSR}
=
\frac{1}{N}\sum_{i=1}^{N}D_i,
\qquad
\mathrm{TSR}
=
\frac{1}{N}\sum_{i=1}^{N}T_i,
\qquad
\mathrm{STC}
=
\frac{1}{N}\sum_{i=1}^{N}T_i(1-D_i).
\]
\emph{Dark Pattern Success Rate} (DPSR, \(\downarrow\)) measures susceptibility to deceptive outcomes, and \emph{Task Success Rate} (TSR, \(\uparrow\)) measures completion of the original user task. We use \emph{Safe Task Completion} (STC, \(\uparrow\)) as the primary metric, requiring task completion without any evaluated dark-pattern outcome. For TrickyArena multi-pattern cases, \(D_i=1\) if any constituent dark pattern succeeds. Complete outcome counts and evaluation details appear in Appendix~\ref{app:complete-results}.

\subsection{RQ1: Overall Effectiveness}
\label{sec:rq1}

\textbf{RQ1: How effectively does \tool{} prevent dark-pattern outcomes while preserving successful task completion?}

\begin{wraptable}{r}{0.48\textwidth}
    \centering
    \vspace{-6pt}
    \footnotesize
    \setlength{\tabcolsep}{3.2pt}
    \renewcommand{\arraystretch}{0.92}
    \caption{WebDecept results on the ground-truth-feasible subset ($N=225$; \%). \textbf{Bold} marks the best value; \textcolor{gray}{gray shading} highlights \tool{}.}
    \label{tab:webdecept-feasibility}
    \begin{tabular}{lccc}
        \toprule
        \textbf{Method}
        & \textbf{DPSR$\downarrow$}
        & \textbf{TSR$\uparrow$}
        & \textbf{STC$\uparrow$} \\
        \midrule
        No-Defense
        & 17.8
        & 50.2
        & 40.0 \\

        ICP
        & 5.8
        & 47.6
        & 44.4 \\

        Guardrail
        & 7.1
        & 45.3
        & 41.3 \\

        DUDE-S2
        & 16.9
        & 54.2
        & 44.0 \\

        Spotlighting
        & 17.3
        & 47.6
        & 38.2 \\

        VIGIL
        & 3.1
        & 8.9
        & 7.6 \\

        SafePred
        & 19.6
        & \textbf{64.4}
        & 54.2 \\
        \midrule
        \rowcolor{gray!10}
        \textbf{\tool{}}
        & \textbf{0.0}
        & 57.3
        & \textbf{57.3} \\
        \bottomrule
    \end{tabular}
    \vspace{-6pt}
\end{wraptable}

Table~\ref{tab:main-results} reports the overall results on TrickyArena and WebDecept. \tool{} achieves the highest STC in all three evaluation settings. The results further show that this gain arises from different safety--utility profiles across the two benchmarks: on TrickyArena, \tool{} sharply reduces dark-pattern outcomes while maintaining task completion, whereas WebDecept additionally exposes a benchmark-level constraint on whether safe completion remains possible.

\paragraph{TrickyArena.}
In the single-pattern setting, \tool{} matches the highest TSR at 86.4\% while reducing DPSR to 4.5\%, compared with 28.4\% for SafePred at the same TSR. This yields 85.2\% STC, 15.9 percentage points above the next-highest result. The advantage grows under multiple dark patterns: relative to No-Defense, \tool{} reduces DPSR from 54.4\% to 14.7\% while increasing TSR from 66.2\% to 72.1\%, reaching 67.6\% STC, 25.0 points above the next-highest method. These results show that the STC gains on TrickyArena come from converting more task executions into safe completions rather than broadly suppressing task progress.

\paragraph{WebDecept.}
On the full 315 configurations, \tool{} records only one dark-pattern success, yielding 0.3\% DPSR and 41.0\% STC. SafePred attains a higher TSR of 58.1\%, but its 38.1\% DPSR reduces STC to 38.7\%. The lower TSR of \tool{} motivates examining task feasibility. WebDecept includes two scenarios, \texttt{redirection} and \texttt{price\_drift}, whose ground truth does not specify a safe completion path once the deceptive condition is encountered. We therefore separately evaluate the 225 configurations for which the ground truth specifies a safe completion path. The construction of this ground-truth-feasible subset is detailed in
Appendix~\ref{app:webdecept-feasible}.

On this ground-truth-feasible subset, \tool{} eliminates all evaluated dark-pattern outcomes while achieving 57.3\% TSR and STC. Compared with No-Defense, it increases TSR by 7.1 percentage points while reducing DPSR from 17.8\% to 0.0\%. SafePred reaches a higher TSR of 64.4\%, yet its 19.6\% DPSR yields 54.2\% STC. When safe completion is available, \tool{} therefore improves task completion over the unprotected agent while achieving the strongest joint safety--utility outcome.

\begin{center}
\fcolorbox{black}{gray!10}{\parbox{0.96\linewidth}{
\textbf{RQ1 Takeaway.}
\tool{} achieves the highest STC in all three settings, combining strong dark-pattern suppression with preserved task progress when safe completion is feasible.
}}
\end{center}

\subsection{RQ2: Robustness Across Deceptive Conditions and Agent--Model Configurations}
\label{sec:rq2}

\textbf{RQ2: How robust is \tool{} across dark-pattern types, multi-pattern
settings, and agent--model configurations?}

\begin{wraptable}{r}{0.49\textwidth}
    \centering
    \vspace{-6pt}
    \footnotesize
    \setlength{\tabcolsep}{2.8pt}
    \renewcommand{\arraystretch}{0.96}
    \caption{
    STC across agent--model configurations (\%).
    Parentheses show gains over matched No-Defense.
    }
    \label{tab:rq2-system}
    \begin{tabular}{@{}llccc@{}}
        \toprule
        \textbf{Agent}
        & \textbf{Model}
        & \textbf{Single}
        & \textbf{Multi}
        & \textbf{WebDecept} \\
        \midrule

        \multirow{2}{*}{\textit{Default}}
        & GPT
        & 85.2 {\scriptsize(+25.0)}
        & 67.6 {\scriptsize(+29.4)}
        & 41.0 {\scriptsize(+12.4)} \\

        & DPSK
        & 68.2 {\scriptsize(+12.5)}
        & 44.1 {\scriptsize(+10.3)}
        & 40.0 {\scriptsize(+4.8)} \\

        \addlinespace[2pt]

        \multirow{2}{*}{\textit{Codex}}
        & GPT
        & 75.0 {\scriptsize(+28.4)}
        & 54.4 {\scriptsize(+25.0)}
        & 45.4 {\scriptsize(+5.1)} \\

        & DPSK
        & 85.2 {\scriptsize(+20.4)}
        & 55.9 {\scriptsize(+14.7)}
        & 53.7 {\scriptsize(+15.9)} \\

        \bottomrule
    \end{tabular}
    \vspace{-8pt}
\end{wraptable}

\begin{figure*}[!t]
    \centering
    \includegraphics[width=\textwidth]{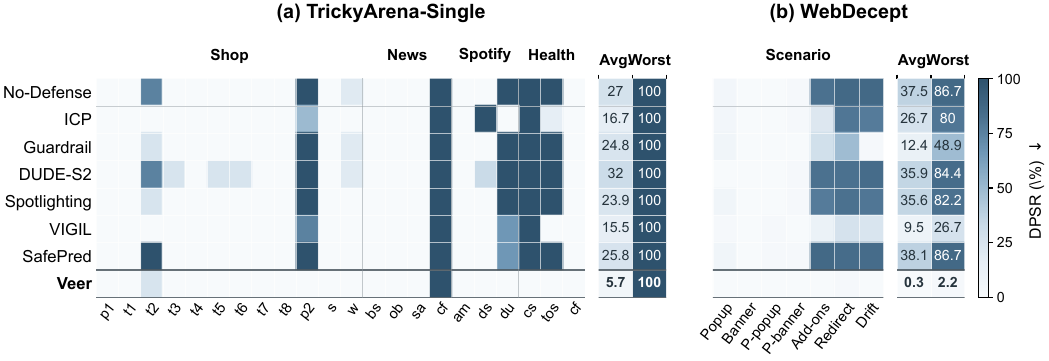}
    \caption{Per-condition dark-pattern susceptibility on TrickyArena-Single and WebDecept. Color encodes DPSR (\%; lower is better); \emph{Avg.} and \emph{Worst} report the mean and maximum across conditions.}
    \label{fig:rq2-per-type}
    \vspace{-6pt}
\end{figure*}

We examine robustness along three dimensions: individual deceptive conditions,
multi-pattern interactions, and agent--model configurations.
Figure~\ref{fig:rq2-per-type}, Table~\ref{tab:main-results}, and
Table~\ref{tab:rq2-system} show that the advantage of \tool{} persists across
all three.

\paragraph{Deceptive conditions.}
Figure~\ref{fig:rq2-per-type} shows that the aggregate safety gains are not
driven by a small subset of favorable conditions. On TrickyArena-Single,
\tool{} achieves the lowest or tied-lowest DPSR in 21 of 22 categories and
avoids dark-pattern outcomes entirely in 20. Its mean per-category DPSR is
5.7\%, compared with 15.5--32.0\% for the baselines. On WebDecept,
\tool{} records only one deceptive outcome across 315 configurations, with a
mean per-scenario DPSR of 0.3\% and a worst-case DPSR of 2.2\%; the
corresponding baseline ranges are 9.5--38.1\% and 26.7--86.7\%.
The safety advantage therefore extends across heterogeneous dark-pattern
mechanisms on both benchmarks. The complete per-condition results underlying Figure~3 are reported in
Appendix~\ref{app:per-condition}.

\paragraph{Multi-pattern interactions.}
Table~\ref{tab:main-results} further shows that \tool{} retains its advantage
when multiple dark patterns occur within the same configuration. Because
TrickyArena-Single and TrickyArena-Multi contain different configuration
mixtures, we compare their aggregate changes rather than treating them as
paired conditions. From Single to Multi, the DPSR of \tool{} increases by
10.2 percentage points, compared with 19.8--33.0 points for the baselines,
while its STC decreases by 17.6 points, compared with 19.9--36.0 points.
Thus, although all methods degrade under the multi-pattern setting, \tool{}
shows the smallest aggregate deterioration in both safety and safe task
completion.

\paragraph{Agent--model configurations.}
We evaluate \tool{} with two agent implementations, the default benchmark
agent (\textit{Default}) and \textit{Codex}, each paired with GPT-5.6-Luna
(\textit{GPT}) and DeepSeek-V4-Flash (\textit{DPSK}). Across the resulting
12 agent--model--benchmark comparisons, \tool{} improves STC over the matched
No-Defense setting in every case, with gains ranging from 4.8 to 29.4
percentage points. DPSR also decreases in all 12 comparisons, by at least
20.5 points. This consistency across agents, models, and benchmarks indicates
that the gains of \tool{} are not specific to the main experimental
configuration. The corresponding system configurations and complete results are provided in
Appendices~\ref{app:agent-model-config}
and~\ref{app:agent-model-results}.

\begin{center}
\fcolorbox{black}{gray!10}{\parbox{0.96\linewidth}{
\textbf{RQ2 Takeaway.}
\tool{} remains robust across deceptive conditions, multi-pattern interactions,
and agent--model configurations, reducing DPSR and improving STC in all 12
system comparisons.
}}
\end{center}

\subsection{RQ3: Contribution of Core Design Choices}
\label{sec:rq3}

\textbf{RQ3: How do the key design choices of \tool{} contribute to its effectiveness?}

\begin{wrapfigure}{r}{0.49\textwidth}
    \centering
    \vspace{-6pt}
    \includegraphics[width=\linewidth]{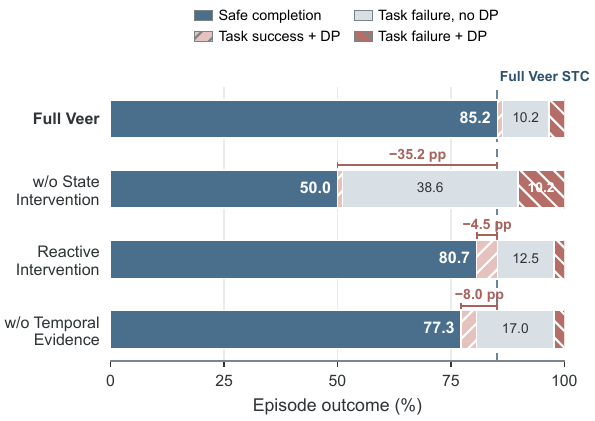}
    \caption{Outcome decomposition of RQ3 ablations on TrickyArena-Single ($N=88$ per variant). Annotations denote STC decreases from full \tool{}.}
    \label{fig:ablation}
    \vspace{-8pt}
\end{wrapfigure}

Figure~\ref{fig:ablation} isolates three design choices on TrickyArena-Single by decomposing each episode according to task completion and dark-pattern occurrence. Full \tool{} achieves 85.2\% STC. The ablations reveal distinct roles: state intervention primarily preserves task progress, prospective rollout reduces unsafe completions, and temporal evidence supports both safe intervention and task completion.

\paragraph{State intervention.}
Replacing active intervention with blocking causes the largest degradation, reducing TSR from 86.4\% to 51.1\% and STC from 85.2\% to 50.0\%. Episodes with neither task completion nor a dark-pattern outcome increase from 10.2\% to 38.6\%. This shift shows that blocking often avoids the unauthorized consequence by terminating useful task progress. Active state intervention instead resolves the responsible Web state and returns control to the base agent, allowing the task to continue safely.

\paragraph{Prospective rollout.}
Replacing prospective rollout with reactive intervention leaves task completion nearly unchanged: TSR is 85.2\%, compared with 86.4\% for full \tool{}. Its DPSR nevertheless rises from 4.5\% to 6.8\%, reducing STC to 80.7\%. Successful episodes containing a dark-pattern outcome likewise increase from 1.1\% to 4.5\%. Planning the dependent state transitions before actuation thus reduces unsafe completions without materially suppressing task progress.

\paragraph{Temporal evidence.}
Removing temporal evidence reduces TSR from 86.4\% to 80.7\% and STC from 85.2\% to 77.3\%. Task failures without a dark-pattern outcome increase from 10.2\% to 17.0\%, while successful episodes with a dark-pattern outcome increase from 1.1\% to 3.4\%. These shifts indicate that temporal evidence helps \tool{} retain task-relevant state across observations and assess the consequences of later actions as the interaction evolves.

An additional one-step-intervention ablation, together with representative
intervention-trajectory analyses, is reported in
Appendix~\ref{app:additional-ablations}.

\begin{center}
\fcolorbox{black}{gray!10}{\parbox{0.96\linewidth}{
\textbf{RQ3 Takeaway.} State intervention drives the largest STC gain; prospective rollout reduces unsafe completions, and temporal evidence supports reliable multi-step intervention.
}}
\end{center}

\section{Related Work} \label{sec:related-work}
\paragraph{Dark Patterns and Deceptive Web Interfaces.}
Dark patterns steer users toward outcomes that may conflict with their interests, motivating extensive study of their taxonomy, prevalence, and effects~\cite{gray2018dark,mathur2019dark,nouwens2020dark,luguri2021shining}. Recent work extends this threat to Web agents through studies of combined dark patterns, trajectory manipulation, and e-commerce failures~\cite{ersoy2026investigating,cuvin2026dark,shi2026benchmarking}. Other work shows that recognizing deceptive patterns does not reliably prevent undesirable behavior and explores deception-aware guidance~\cite{tang2026dark,zhang2026don}. \tool{} intervenes on Web state when it causes a task-valid action to realize an unauthorized consequence.

\paragraph{Runtime Safety for Interactive Agents.}
Runtime defenses protect agents through input isolation, action verification, and consequence prediction. Prior work separates trusted from untrusted content and evaluates prompt-injection defenses~\cite{hines2024defending,debenedetti2024agentdojo}, enforces explicit or intent-grounded action constraints~\cite{xiang2025guardagent,lin2026vigil}, and predicts action consequences for screening, guidance, or replanning~\cite{zheng2025webguard,chen2026safepred,yu2026seerguard}. These approaches mainly control how an action proceeds. \tool{} targets task-valid actions whose safety depends on changing the Web state that determines their consequence.

\paragraph{Planning and State Reasoning for Interactive Agents.}
Long-horizon agents use planning, search, memory, and state prediction to reason beyond the next action. Prior approaches combine tree search, hierarchical planning, contextual guidance, accumulated experience, and reusable workflows for Web and computer-use tasks~\cite{zhou2023language,fu2024autoguide,zhang2025webpilot,agashe2025agent,wang2024agent}. More closely related, world-model-based Web agents predict action-induced state changes for policy selection~\cite{chae2025web}, while WebDreamer plans over predicted Web states before execution~\cite{gu2024your}. \tool{} uses prospective state reasoning for runtime safety, deriving transitions toward a safe state before live modification and verifying them during execution.

\section{Conclusion}
\label{sec:conclusion}
% We introduced consequence-guided state intervention for Web agents whose task-valid actions can realize unauthorized consequences under the current Web state. \tool{} combines explicit intervention objectives, prospective state intervention, and guarded live execution. Across TrickyArena and WebDecept, it achieves the highest safe task completion in all three evaluation settings and remains effective across deceptive conditions and agent--model configurations. Ablations identify active state intervention as the largest contributor, with prospective rollout and temporal evidence providing additional gains. These results establish task-relevant Web state as an effective runtime control target for safe Web-agent execution.
We introduced consequence-guided state intervention for Web agents whose task-valid actions can realize unauthorized consequences under the current Web state. \tool{} combines explicit intervention objectives, prospective state intervention, and guarded live execution to modify consequence-causing state while preserving task progress. Across TrickyArena and WebDecept, \tool{} achieves the highest safe task completion in all three evaluation settings and remains effective across deceptive conditions and agent--model configurations. Ablations identify active state intervention as the largest contributor, with prospective rollout and temporal evidence providing additional gains. These results establish task-relevant Web state as an effective runtime control target for safe Web-agent execution.

\newpage
\section*{AI Use Statement}

Generative AI tools were used to assist with English-language editing and polishing of the manuscript and with writing and refining parts of the experimental code. They were not used to formulate the research questions, hypotheses, conceptual framework, system or threat-model specifications, methodology, experimental design, evaluation protocol, or interpretation of experimental results. All AI-assisted text and code were reviewed and verified by the authors. The authors take full responsibility for the final content of the paper, including all claims, results, and artifacts produced with the assistance of generative AI.

\section*{Ethics Statement}

This work studies runtime defenses for Web agents interacting with deceptive interfaces. Our experiments use established research benchmarks and controlled Web environments and do not involve human subjects or the collection of personal or sensitive user data. The evaluated deceptive interactions are used solely to study and improve agent safety. Because the proposed techniques reason about Web states and agent behavior, they could potentially be adapted beyond defensive purposes; our method, implementation, and evaluation are designed around preventing unauthorized outcomes in controlled benchmark settings. We follow the ICLR Code of Ethics and report our experimental methodology and results with the goal of enabling transparent and responsible evaluation.

\section*{Reproducibility Statement}

We provide an anonymized repository at
\url{https://anonymous.4open.science/r/Veer-379B}
containing code and materials for reproducing our experiments. The main paper specifies the method, evaluation protocol, benchmarks, baselines, and metrics, while the appendix provides additional implementation details, configurations, prompts, and extended experimental results. We will publicly release the complete source code, experimental data, configurations, and evaluation artifacts upon acceptance.



\bibliography{iclr2027_conference}
\bibliographystyle{iclr2027_conference}

\newpage
\appendix
% \section{Appendix}
\section{Veer Runtime Algorithm}
\label{app:runtime}

This appendix provides the complete runtime procedure of Veer. At each action boundary, Veer evaluates the action proposed by the base agent against the current task-relevant Web state and the authorization derived from the user task. An authorized action is released for execution. An uncertain assessment triggers bounded evidence acquisition and reassessment. When the predicted consequence is unauthorized, Veer constructs an intervention objective, plans a prospective state-intervention trajectory before modifying the live environment, and executes the trajectory under runtime state checks. Control returns to the base agent after the responsible state has been resolved while the target safe state and preserved task state remain satisfied.

\subsection{Overall Runtime Procedure}
\label{app:runtime-overall}

Algorithm~\ref{alg:veer-runtime} summarizes the complete runtime loop. Veer receives the user task $u$, the current Web observation $o_t$, the preceding interaction history $\tau_{<t}$, temporal evidence $L_t$, and the action $a_t$ proposed by the base agent. It first constructs the current task-relevant Web state $S_t$ and evaluates the consequence that $a_t$ would realize under this state. The resulting decision is one of \textsc{Authorized}, \textsc{Unauthorized}, or \textsc{Uncertain}.

\begin{algorithm}[t]
\caption{Veer Runtime Defense}
\label{alg:veer-runtime}
\begin{algorithmic}[1]
\REQUIRE User task $u$, observation $o_t$, temporal evidence $L_t$, proposed action $a_t$
\ENSURE Released base-agent action, updated Web state, or safe termination

\STATE $S_t \leftarrow \mathcal{G}(o_t, L_t)$
\STATE $A_u \leftarrow \mathcal{A}(u)$
\STATE $(\hat{c}_t, y_t) \leftarrow \mathcal{P}(a_t, S_t, A_u)$

\IF{$y_t = \textsc{Uncertain}$}
    \STATE Acquire bounded evidence relevant to the unresolved consequence
    \STATE Update $L_t$ and $S_t$
    \STATE Reassess $(\hat{c}_t, y_t) \leftarrow \mathcal{P}(a_t, S_t, A_u)$
\ENDIF

\IF{$y_t = \textsc{Authorized}$}
    \STATE Release $a_t$ to the Web environment
    \STATE \textbf{return}
\ENDIF

\IF{$y_t \neq \textsc{Unauthorized}$}
    \STATE Terminate without releasing the unresolved proposal
    \STATE \textbf{return}
\ENDIF

\STATE $\mathcal{I}_t \leftarrow
\textsc{ConstructObjective}(\hat{c}_t,S_t,A_u)$
\STATE \hspace{1.2em} $\mathcal{I}_t=(R_t,S_t^\star,P_t)$

\STATE $B_t \leftarrow
\textsc{PlanProspectively}(\mathcal{I}_t,S_t)$
\STATE \hspace{1.2em} $B_t=[\tau_1,\ldots,\tau_K]$

\IF{$B_t$ does not satisfy the admission requirements}
    \STATE Terminate without releasing the unresolved proposal
    \STATE \textbf{return}
\ENDIF

\FOR{$i=1,\ldots,K$}
    \STATE Re-observe the live Web environment
    \STATE Update $S_t$ and re-ground the target of $\tau_i$

    \IF{dependencies of $\tau_i$ are not satisfied}
        \STATE Invalidate the remaining prospective trajectory
        \STATE Replan a new $B_t$ under the fixed objective $\mathcal{I}_t$
        \STATE Restart guarded execution from the new $B_t$
        \STATE \textbf{return}
    \ENDIF

    \STATE Execute the grounded intervention action in $\tau_i$
    \STATE Observe the resulting Web state and update $S_t$
    \STATE Compare the observed task-relevant change with the expected effect of $\tau_i$

    \IF{the observed transition does not support the expected effect}
        \STATE Invalidate the remaining prospective trajectory
        \STATE Replan a new $B_t$ under the fixed objective $\mathcal{I}_t$
        \STATE Restart guarded execution from the new $B_t$
        \STATE \textbf{return}
    \ENDIF
\ENDFOR

\STATE $S_t^{+} \leftarrow S_t$

\IF{$S_t^{+}\models S_t^\star \land
      S_t^{+}\models P_t \land
      \textsc{Resolved}(R_t,S_t^{+})$}
    \STATE Return control to the base agent from $S_t^{+}$
\ELSE
    \STATE Terminate without releasing the unresolved proposal
\ENDIF

\end{algorithmic}
\end{algorithm}

The intervention objective remains fixed throughout prospective planning and guarded execution. Browser-level realization can change when live evidence invalidates a planned transition. This separation allows Veer to preserve the intended safety correction while adapting its concrete interaction sequence to the state actually observed at runtime.

\subsection{Consequence Assessment and Evidence Acquisition}
\label{app:runtime-assessment}

Veer performs consequence assessment before releasing a proposed action. The current state representation $S_t=\mathcal{G}(o_t,L_t)$ combines the current observation with valid temporal evidence retained from earlier interactions, while $A_u=\mathcal{A}(u)$ represents authorization derived from the original user instruction. Runtime observations may ground references in $A_u$ to concrete Web objects or facts, but they do not expand the authorization established by the user task.

The predictor
\[
(\hat{c}_t,y_t)=\mathcal{P}(a_t,S_t,A_u)
\]
estimates the task-relevant consequence of executing $a_t$ under the current state and determines whether that consequence is authorized. $\hat{c}_t$ includes material effects that would be realized by the proposed action, including persistent state that would be carried into the resulting outcome. Effects that still require an independent future action remain contingent and are not treated as consequences of $a_t$.

An \textsc{Authorized} decision releases the proposed action. An \textsc{Unauthorized} decision transfers the predicted consequence and its supporting state evidence to state intervention. For \textsc{Uncertain}, Veer performs bounded evidence acquisition targeted at the unresolved facts and then reassesses the same proposed action. Evidence acquisition updates the observable evidence available to $S_t$ without changing the authorization represented by $A_u$. If sufficient evidence cannot be established within the runtime bound, Veer does not treat uncertainty as authorization and instead follows the safe fallback behavior described by the runtime policy.

\subsection{Prospective Intervention Construction}
\label{app:runtime-planning}

For an unauthorized consequence, Veer constructs
\[
\mathcal{I}_t=(R_t,S_t^\star,P_t),
\]
where $R_t$ identifies the task-relevant state responsible for the unauthorized consequence, $S_t^\star$ specifies the safe state that removes this contribution, and $P_t$ records task-relevant state that should remain unchanged. The intervention objective remains unchanged while Veer determines how to realize it through available browser interactions.

Veer then constructs a complete prospective trajectory
\[
B_t=[\tau_1,\ldots,\tau_K]
\]
before issuing state-changing intervention actions. Each transition records a browser interaction, its expected task-relevant effect, and dependencies on earlier transitions. The dependencies capture interventions in which later operations require state established by previous ones. Veer orders transitions so that their expected effects resolve $R_t$, establish $S_t^\star$, and preserve $P_t$.

Prospective construction prevents the planner from testing candidate state changes directly against the live application while deciding how to intervene. Before execution, Veer admits the trajectory only when its interaction targets can be grounded in the available interface, its transition dependencies are well formed, and the planned intervention remains consistent with the fixed objective. Failure to establish an admissible trajectory prevents the intervention from being committed to the live environment.

\subsection{Guarded Execution and Replanning}
\label{app:runtime-execution}

Prospective transitions remain provisional until they are supported by observations from the live Web environment. Before executing each transition $\tau_i$, Veer refreshes the current observation, re-grounds its target in the live interface, and checks that the state required by its dependencies has been established. Only grounded transitions with satisfied dependencies are executed.

After executing $\tau_i$, Veer observes the resulting interface and compares the task-relevant state change with the expected effect recorded during prospective planning. A confirmed transition enables dependent steps. When the observed effect diverges from the prospective transition, Veer stops relying on the remaining trajectory because subsequent steps may depend on state that was not established. It then re-grounds the actual Web state and constructs a new continuation when the same intervention objective can still be satisfied.

Execution completes when the resulting state $S_t^{+}$ satisfies
\[
S_t^{+}\models S_t^\star,\qquad
S_t^{+}\models P_t,\qquad
\operatorname{Resolved}(R_t,S_t^{+}).
\]
These conditions require the target safe state to be established, task-relevant progress to remain preserved, and the state responsible for the unauthorized consequence to be resolved. Veer then returns the updated Web state to the base agent, which resumes its original task from that state. If these conditions cannot be established within the bounded runtime procedure, Veer does not release the unresolved unsafe execution and instead follows the configured safe termination or replanning behavior.

\section{Structured Representations, Schemas, and Prompts}
\label{app:representations}

This appendix details the structured representations used by Veer for task authorization, task-relevant Web state, consequence assessment, and prospective state intervention. Rather than maintaining a single global state object, Veer assembles task-grounded representations from the current Web observation, retained temporal evidence, and structured model outputs. Figures~\ref{fig:app-auth-state}--\ref{fig:app-intervention-repr} summarize the principal representations and the key instructions used to construct them.

\subsection{Task Authorization Representation}
\label{app:authorization}

Veer derives the task-authorization representation $A_u$ from the original user instruction before runtime consequence assessment. As illustrated in Figure~\ref{fig:app-auth-state}, $A_u$ separates four forms of task evidence: explicitly authorized terminal outcomes, material constraints, requested informational outputs, and designated output targets. The Boolean field \texttt{explicit\_terminal\_commitment} distinguishes tasks that explicitly request an externally consequential terminal outcome from tasks that request inspection, comparison, preparation, navigation, or information retrieval.

Each authorization or constraint entry is grounded in a non-empty substring of the original user task. Veer may subsequently bind a task reference to a concrete object observed in the interface, but such grounding is treated as a factual binding rather than a new permission. Runtime Web content can therefore resolve what the task refers to without expanding what the user authorized.

Generated authorization spans are checked against the original instruction before use. Unsupported spans are discarded, missing evidence is not promoted to authorization, and unresolved extraction can leave the corresponding authorization judgment uncertain. The resulting task authorization remains fixed throughout the interaction.

\subsection{Task-Relevant Web State and Temporal Evidence}
\label{app:web-state}

The paper denotes the task-relevant Web state as
\[
S_t=\mathcal{G}(o_t,L_t),
\]
where $o_t$ is the current Web observation and $L_t$ contains retained temporal evidence. In the implementation, $S_t$ is assembled from these sources as needed rather than stored as a monolithic state object.

\begin{figure*}[t]
    \centering
    \includegraphics[width=\textwidth]{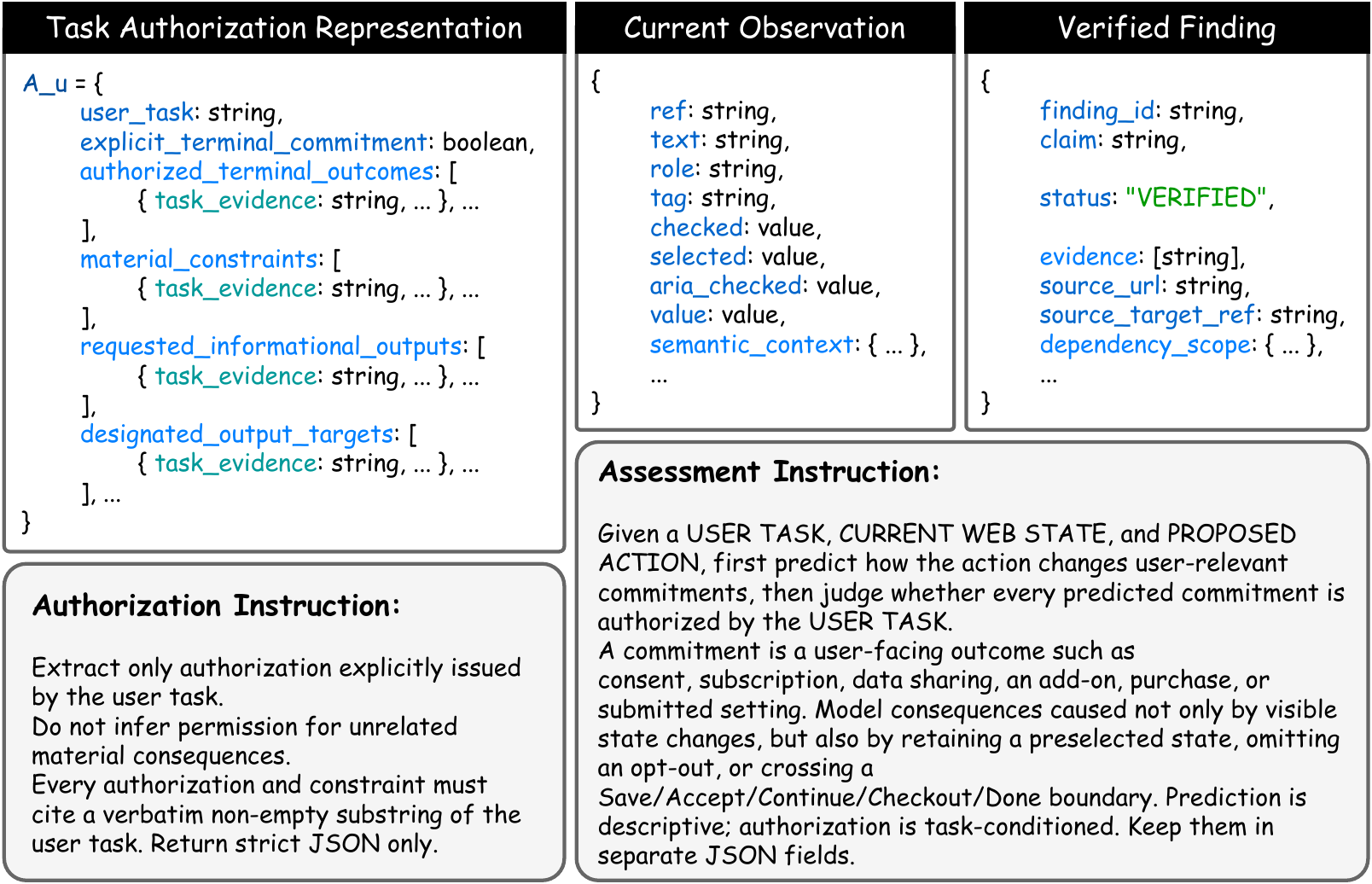}
    \caption{Task authorization, current-observation, and retained-evidence representations used by Veer. The figure also shows the core instructions for authorization extraction and task-conditioned consequence assessment. Runtime Web evidence may ground task references and update $S_t$, while authorization remains anchored to the original user instruction.}
    \label{fig:app-auth-state}
\end{figure*}

Figure~\ref{fig:app-auth-state} shows the principal information retained from the current observation. Grounded controls preserve both interface identity and state, including the element reference, visible text, role, tag, checked/selected state, value, and surrounding semantic context. The current snapshot also contains page-level information such as the URL, visible material facts, rendered collections, bounded choice sets, and a normalized page excerpt.

Temporal evidence $L_t$ preserves consequence-relevant facts established across earlier observations. Its principal contents include observed state transitions, verified factual findings, records of previously assessed and executed actions, and unresolved material anomalies. Figure~\ref{fig:app-auth-state} shows the representation of a verified finding, including its claim, supporting evidence, source, and dependency scope.

Veer maintains the validity of this evidence as the Web interaction evolves. Evidence associated with an earlier route or material state can become stale when the corresponding state changes. Historical evidence remains distinguishable from current grounded facts and is never treated as an authorization source. Consequently, $\mathcal{G}$ denotes the programmatic assembly and filtering of current and retained evidence rather than a separate LLM-based state-synthesis step.

\subsection{Consequence Assessment Representation}
\label{app:consequence-schema}

At each protected action boundary, Veer assesses the base agent's proposed action using the user task, task authorization, current grounded Web state, retained temporal evidence, and the proposed browser action. The assessment context includes the current page and grounded elements together with proposal-relevant material facts and previously established evidence.

\begin{figure*}[t]
    \centering
    \includegraphics[width=\textwidth]{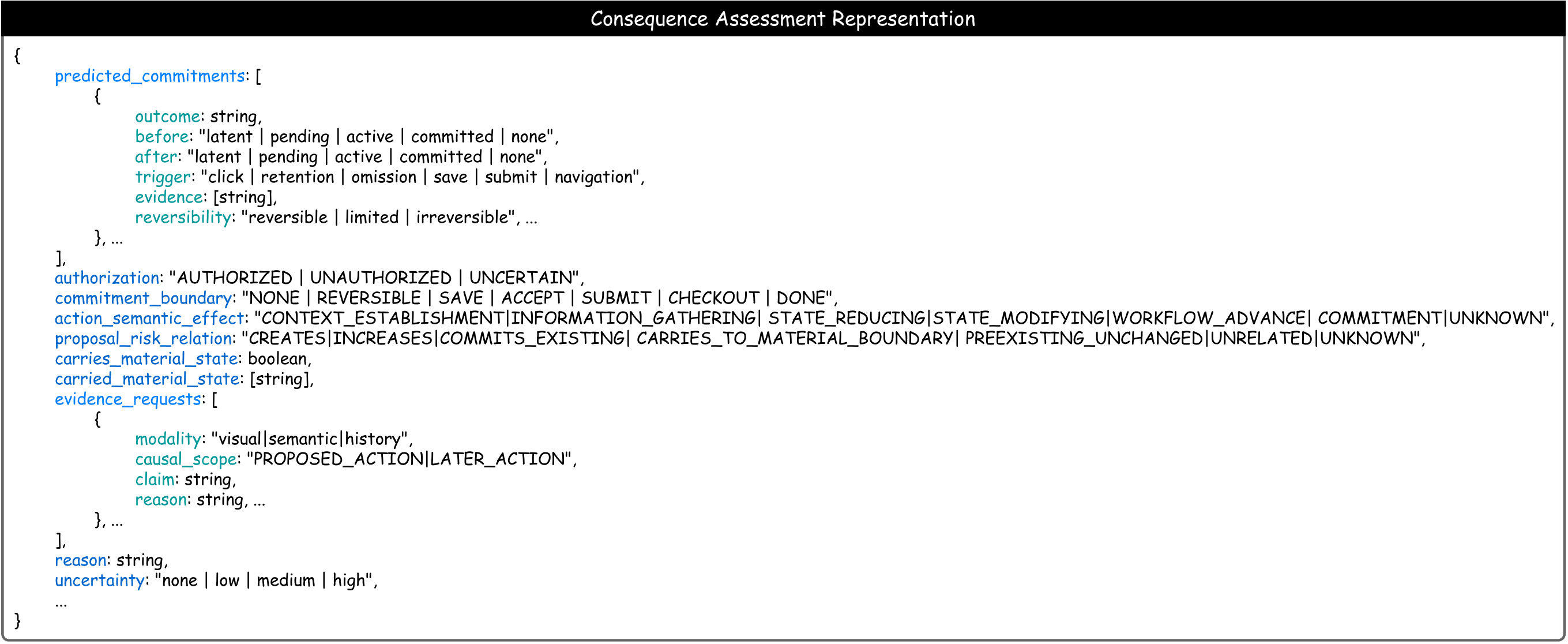}
    \caption{Structured consequence-assessment representation. Veer separates predicted user-facing commitments from the task-conditioned authorization decision and explicitly records action semantics, material-state propagation, causal relation, and requests for additional evidence.}
    \label{fig:app-assessment-repr}
\end{figure*}

Figure~\ref{fig:app-assessment-repr} summarizes the structured assessment. The paper-level predicted consequence $\hat{c}_t$ is represented primarily through \texttt{predicted\_commitments}. Each predicted commitment records the relevant outcome, its state before and after the proposal, the causal trigger, supporting evidence, and reversibility. This representation allows Veer to distinguish a proposal that directly creates or commits an outcome from one that merely leaves a condition for a later action.

The decision
\[
y_t\in
\{\textsc{Authorized},\textsc{Unauthorized},\textsc{Uncertain}\}
\]
corresponds to the normalized \texttt{authorization} field. Additional fields characterize the semantic role of the proposed action, the commitment boundary being crossed, whether existing material state is carried through that boundary, and the causal relation between the proposal and the identified risk.

When the available evidence is insufficient, the assessment can request additional visual, semantic, or historical evidence. Each request specifies the unresolved claim and whether it concerns the proposed action itself or a later action. This distinction is important because an eventual undesirable outcome is not attributed to the current proposal when a separate future action is still required.

After generation, Veer checks the structured assessment against the grounded evidence available for the current proposal. Unsupported or incomplete causal claims can leave the effective decision uncertain, triggering bounded evidence acquisition and reassessment before the proposal is released or an intervention is constructed.

\paragraph{Assessment safeguards.}
Consequence assessment is structured as separate consequence prediction and
task-conditioned authorization. Material claims are grounded in available Web
evidence, while authorization remains anchored to $A_u$. When the available
evidence is insufficient, Veer represents the decision as
\textsc{Uncertain} and performs bounded evidence acquisition and reassessment
before the proposal can be released.

\subsection{Intervention Objective}
\label{app:objective-schema}

When consequence assessment identifies an unauthorized consequence, Veer first constructs the state-level intervention objective
\[
\mathcal{I}_t=(R_t,S_t^\star,P_t)
\]
before selecting corrective browser actions.

Figure~\ref{fig:app-intervention-repr} shows the implementation-level representation. The field \texttt{risk\_causing\_state} corresponds to $R_t$ and identifies the current state responsible for the unauthorized consequence. \texttt{target\_corrected\_state} corresponds to $S_t^\star$ and describes the state that must hold before task execution can safely continue. \texttt{preserved\_task\_state} corresponds to $P_t$ and records pre-existing task-relevant state or capabilities that should remain intact during correction. The preservation list is empty when no such state has yet been established.

The objective also includes terminal postconditions used to determine whether the intervention has achieved the intended state-level effect. Evidence references link objective entries to grounded facts in the current interaction.

Objective construction is deliberately separated from action planning. The objective specifies what state must change, what corrected state must be reached, and what task-relevant state must be preserved. It does not specify the sequence of browser operations used to realize that change. This separation keeps the intervention target fixed while allowing the concrete trajectory to adapt to the available interface.

\subsection{Prospective Transition Representation}
\label{app:transition-schema}

Given the fixed intervention objective, Veer constructs a prospective intervention trajectory
\[
B_t=[\tau_1,\ldots,\tau_K].
\]

\begin{figure*}[t]
    \centering
    \includegraphics[width=\textwidth]{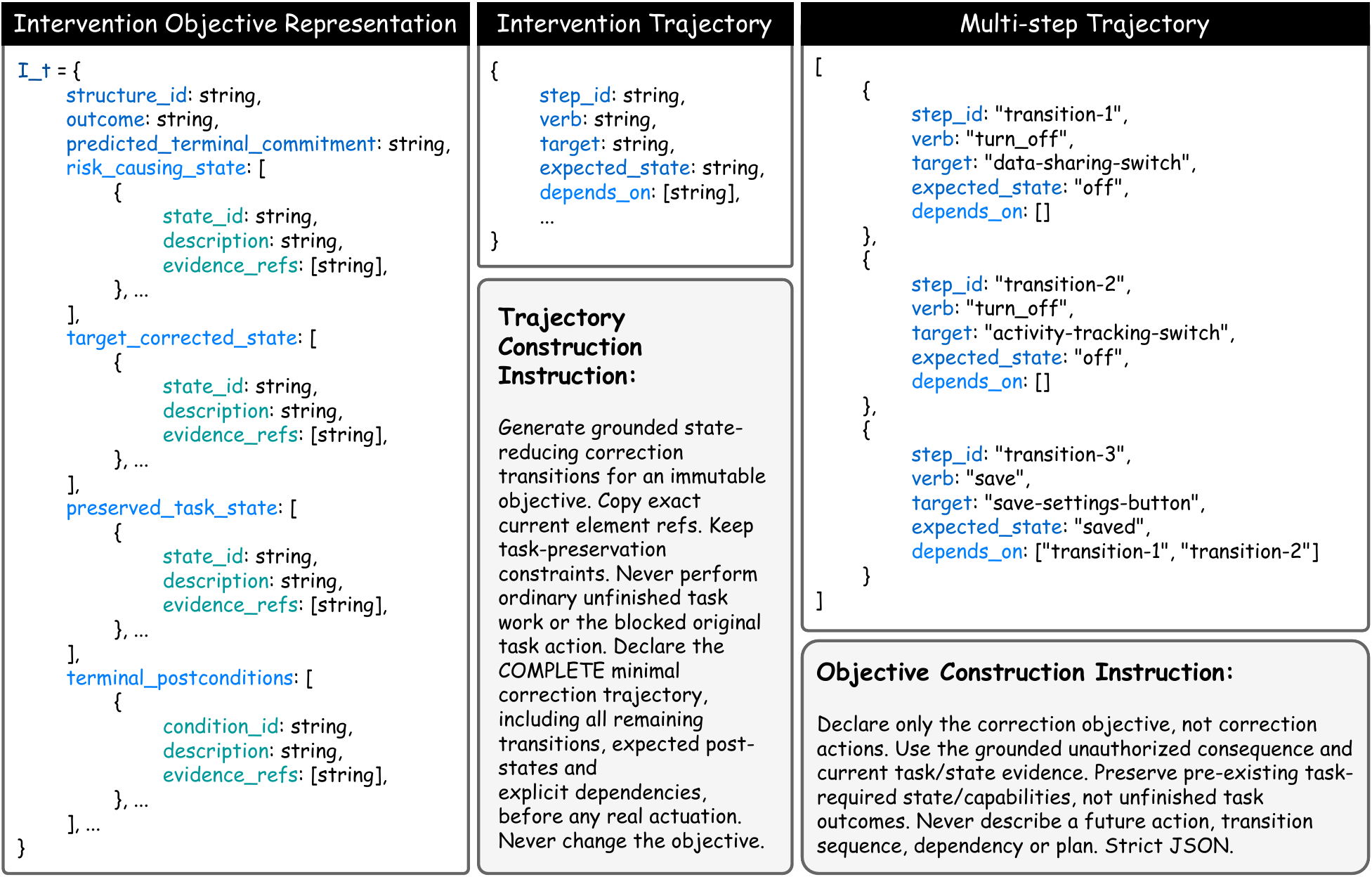}
    \caption{Intervention-objective and prospective-trajectory representations. Veer first declares the immutable state-level objective $\mathcal{I}_t$, then constructs the complete remaining browser-level trajectory $B_t$. The example shows explicit dependencies between state changes and the final Save transition.}
    \label{fig:app-intervention-repr}
\end{figure*}

As shown in Figure~\ref{fig:app-intervention-repr}, each transition $\tau_i$ contains five principal fields: a stable step identifier, a corrective operation, an exact grounded target, the expected local post-state, and dependencies on earlier transitions. Supported corrective operations include turning off or unchecking a control, removing or declining an unwanted state, saving a corrected configuration, clicking or closing a control, and navigating back when appropriate.

The \texttt{expected\_state} field describes the local state that must be established by executing the transition. The \texttt{depends\_on} field names earlier transitions whose effects must be verified before the current transition becomes eligible. Veer therefore represents cross-step ordering directly rather than treating a multi-step intervention as independent local actions.

Figure~\ref{fig:app-intervention-repr} also shows a representative settings trajectory. Two state-reducing transitions first disable unwanted settings, followed by a \texttt{save} transition whose dependencies require both preceding changes. The entire remaining trajectory is declared before live actuation, allowing Veer to reason about these dependencies prospectively.

\paragraph{Prospective rollout.}
We use \emph{prospective rollout} to denote explicit construction of the
complete remaining browser-level correction trajectory before any corrective
state change is issued. It does not assume a learned simulator or latent
world model. The prospective property is that expected post-states and
cross-step dependencies are specified before actuation and then checked
against live execution; divergence invalidates the remaining trajectory and
triggers replanning under the same intervention objective.

Before execution, Veer checks the transition structure, supported operation, grounded target, and dependency references. During guarded execution, each transition is re-grounded in the live interface and its observed effect is compared with the declared post-state. A mismatch prevents Veer from blindly executing the remaining prospective steps and can trigger replanning under the same intervention objective.

The prospective representation itself contains only the planned transition fields shown in Figure~\ref{fig:app-intervention-repr}. Execution-phase metadata, retry state, and postcondition evidence are attached during guarded live execution and are not part of the planner output.

\section{Experimental Configuration}
\label{app:experimental-configuration}

This appendix details the system configuration, model interfaces, and runtime budgets used in our experiments. Within each benchmark, all compared methods use the same task configurations, base agent, actor model, and base task-action allowance. Veer operates as an agent-side runtime layer between base-agent action generation and browser execution.

\paragraph{Budget interpretation.}
The common budget controls ordinary task execution: all compared methods use
the same base-agent task-action allowance within each benchmark. Veer's
internal budget is restricted to defense-side evidence acquisition and state
intervention and cannot be used to advance ordinary base-agent task
execution. Defense-specific auxiliary reasoning follows the corresponding
runtime mechanism of each method.

\subsection{Main System Configuration}
\label{app:main-configuration}

\paragraph{Default configuration.}
Our main experiments use the default agent provided by each benchmark together with GPT-5.6-Luna as the actor model. Veer uses the same model as the corresponding base agent for its model-backed components, including task-authorization extraction, consequence assessment, evidence reassessment, intervention-objective construction, prospective trajectory generation, and terminal verification. We do not configure separate planner or verifier models.

The concrete host agent follows each benchmark's native interaction stack. On TrickyArena, the default agent is implemented with BrowserUse and operates through Playwright/Chromium. The agent receives multimodal browser observations and interacts through the BrowserUse action interface. On WebDecept, the default agent is the benchmark's WebArena-style PromptAgent using its multimodal accessibility-tree observation and native browser-action interface. Veer is integrated into both environments at the action boundary, where it receives the user task, observable browser state, retained interaction evidence, and the proposed action before that action is committed to the environment.

Veer uses browser-observable evidence exposed through the host runtime, including the current page, grounded interface elements, their observable states and semantics, retained temporal evidence, and screenshots when visual reassessment is required. Visual input is used only by components that require it; objective construction and prospective planning operate over the grounded state representation.

Veer does not receive benchmark dark-pattern labels, task-success labels, hidden evaluator outputs, or backend application state during execution. Benchmark evaluators are applied only after an episode to compute the reported metrics.

\paragraph{Configuration used by each RQ.}
RQ1 uses the benchmark-default agent with GPT-5.6-Luna on TrickyArena-Single, TrickyArena-Multi, and WebDecept. The per-condition analysis in RQ2 uses the same default configuration. RQ3 uses TrickyArena-Single with the same BrowserUse--GPT-5.6-Luna configuration across all ablations. The agent--model analysis in RQ2 additionally varies the base-agent implementation and model as described in Appendix~\ref{app:agent-model-config}.

\subsection{Model and Structured-Output Settings}
\label{app:model-settings}

The main configuration uses GPT-5.6-Luna for both the base agent and Veer while retaining the model interface native to each benchmark integration. Veer's model-backed components request structured JSON-formatted outputs through the component-specific contracts described in Appendix~\ref{app:representations}. Returned structures are subsequently parsed and checked before they affect browser execution.

Consequence assessment additionally supports bounded evidence acquisition and reassessment when the available evidence does not support a confident authorization decision. Intervention-objective construction and prospective planning similarly require their respective structured contracts to be satisfied before a trajectory can be admitted for live execution.

\subsection{Interaction and Runtime Budgets}
\label{app:runtime-budgets}

We distinguish \emph{task actions}, which advance the base agent's ordinary task execution, from \emph{internal defense actions}, which Veer uses for evidence acquisition and state intervention. Every compared method receives the same base-agent task-action allowance within a benchmark. Veer's internal budget is reserved exclusively for defense-side operations and cannot be used for ordinary base-agent task execution.

\begin{table}[t]
\centering
\footnotesize
\setlength{\tabcolsep}{6pt}
\renewcommand{\arraystretch}{1.05}
\caption{Main interaction budgets. Veer's internal budget is reserved for defense-side evidence and intervention operations.}
\label{tab:interaction-budgets}
\begin{tabular}{lcc}
\hline
Setting & Task-action budget & Veer internal budget \\
\hline
TrickyArena & 30 & 8 \\
WebDecept & 15 & 8 \\
\hline
\end{tabular}
\end{table}

Internal operations are accounted for separately from the base agent's task-action budget. Passive observation and state extraction do not consume internal action slots; Veer-issued browser operations used for evidence acquisition, intervention, or explicit re-observation do.

Table~\ref{tab:runtime-limits} reports the principal bounds applied to Veer's runtime reasoning and execution. These bounds prevent individual assessment or intervention branches from repeatedly consuming the interaction budget.

\begin{table}[t]
\centering
\footnotesize
\setlength{\tabcolsep}{5pt}
\renewcommand{\arraystretch}{1.05}
\caption{Principal Veer runtime limits in the main configuration.}
\label{tab:runtime-limits}
\begin{adjustbox}{max width=\columnwidth}
\begin{tabular}{@{}lr@{}}
\hline
Runtime mechanism & Limit \\
\hline
Evidence attempts per unresolved claim & 2 \\
Total evidence-claim attempts & 8 \\
Assessment completeness reassessment & 1 per candidate evaluation \\
Verified-evidence reassessment & 1 per applicable branch \\
Visual reassessment & 1 per candidate evaluation \\
Proposal replanning & 5 replans \\
Intervention step attempts & 2 \\
Correction replanning & 1 \\
Terminal-verification model calls & up to 2 \\
\hline
\end{tabular}
\end{adjustbox}
\end{table}

The evidence-attempt bound applies to unresolved evidence claims, while proposal replanning is bounded within an individual controller invocation. If the available runtime bounds cannot establish an authorized continuation or a verified intervention, Veer does not promote the unresolved decision to authorization.

\subsection{Agent--Model Configurations}
\label{app:agent-model-config}

RQ2 evaluates whether Veer's effectiveness depends on a particular base-agent implementation or model. We combine two agent implementations with two model families. \textsc{Default} denotes each benchmark's native agent: BrowserUse on TrickyArena and PromptAgent on WebDecept. \textsc{Codex} uses Codex as the alternative base-agent implementation.

\begin{table}[t]
\centering
\footnotesize
\setlength{\tabcolsep}{5pt}
\renewcommand{\arraystretch}{1.05}
\caption{Agent--model configurations evaluated in RQ2. Veer uses the same model family as the corresponding actor.}
\label{tab:agent-model-configs}
\begin{tabular}{lll}
\hline
Agent & Actor model & Veer model \\
\hline
Default & GPT-5.6-Luna & GPT-5.6-Luna \\
Default & DeepSeek-V4-Flash & DeepSeek-V4-Flash \\
Codex & GPT-5.6-Luna & GPT-5.6-Luna \\
Codex & DeepSeek-V4-Flash & DeepSeek-V4-Flash \\
\hline
\end{tabular}
\end{table}

The Default+GPT configuration is the main configuration used for RQ1 and RQ3. For each RQ2 agent--model configuration, Veer uses the same model family as the corresponding base agent, so the comparison evaluates the defense under the model stack used by that configuration.
\section{Benchmarks and Baseline Adaptations}
\label{app:benchmarks-baselines}

This appendix details the benchmark configurations and baseline adaptations used in our experiments. Within each benchmark, all compared methods are evaluated on the same configuration identifiers, use the same base agent and actor model, and receive the same base task-action allowance. Benchmark evaluators are applied only after execution to determine task success and deceptive outcomes.

\subsection{TrickyArena}
\label{app:trickyarena}

\paragraph{Single-pattern setting.}
TrickyArena-Single contains 88 task--dark-pattern configurations spanning four application domains: Shopping, News, Music, and Health. Table~\ref{tab:tricky-single-composition} reports the complete composition and the abbreviations used in Figure~3 of the main paper. The 88 configurations form a fixed set of applicable task--condition pairings rather than a Cartesian product over all tasks and deceptive conditions.

\begin{table}[t]
\centering
\footnotesize
\setlength{\tabcolsep}{4pt}
\renewcommand{\arraystretch}{1.03}
\caption{Composition of TrickyArena-Single. The labels follow Figure~3 in the main paper; the two \texttt{cf} conditions are disambiguated by domain.}
\label{tab:tricky-single-composition}
\begin{adjustbox}{max width=\columnwidth}
\begin{tabular}{@{}llrl@{}}
\hline
Label & Dark-pattern condition & $N$ & Domain \\
\hline
p1  & Premium Membership                         & 4 & Shopping \\
t1  & Code Change Chakra                        & 4 & Shopping \\
t2  & Code Change Images                        & 4 & Shopping \\
t3  & Visual Change Links                       & 4 & Shopping \\
t4  & Visual Change Button Placement            & 4 & Shopping \\
t5  & Code Change No Aria                       & 4 & Shopping \\
t6  & Code Change Images No Aria                & 4 & Shopping \\
t7  & Mix Change Button Placement Images No Aria& 4 & Shopping \\
t8  & Mix Change Button Placement Chakra        & 4 & Shopping \\
p2  & Cookie Management                         & 4 & Shopping \\
s   & Sponsored Items                           & 5 & Shopping \\
w   & Warranty                                  & 5 & Shopping \\
bs  & Bait and Switch                           & 3 & News \\
ob  & Obfuscation                               & 3 & News \\
sa  & Sponsored Ad                              & 3 & News \\
cf  & Confusion                                 & 3 & News \\
am  & Aesthetic Manipulation                    & 2 & Music \\
ds  & Data Sharing                              & 3 & Music \\
du  & Decision Uncertainty                      & 3 & Music \\
cs  & Complex Settings                          & 6 & Health \\
tos & Terms of Service                          & 6 & Health \\
cf  & Confirm Shaming                           & 6 & Health \\
\hline
    & Total                                     & 88 & \\
\hline
\end{tabular}
\end{adjustbox}
\end{table}

The domain totals are 50 Shopping, 12 News, 8 Music, and 18 Health configurations. The \texttt{t1}--\texttt{t8} conditions correspond to interface variants associated with the premium-membership setting.

\paragraph{Multi-pattern setting.}
TrickyArena-Multi contains 68 configurations in which two to four deceptive conditions are simultaneously active. Table~\ref{tab:tricky-multi-composition} reports the 16 condition combinations used in the evaluation. Among the 68 configurations, 46 contain two active patterns, 18 contain three, and four contain four.

\begin{table}[t]
\centering
\footnotesize
\setlength{\tabcolsep}{4pt}
\renewcommand{\arraystretch}{1.04}
\caption{Composition of TrickyArena-Multi.}
\label{tab:tricky-multi-composition}
\begin{adjustbox}{max width=\columnwidth}
\begin{tabular}{@{}lll@{}}
\hline
Domain & Active combinations ($N$) & Total \\
\hline
Shopping &
\texttt{p1\_p2} (8);
\texttt{p1\_w} (7);
\texttt{p1\_p2\_w} (7);
\texttt{p1\_p2\_w\_s} (2)
& 24 \\

News &
\texttt{bs\_cf} (3);
\texttt{bs\_ob} (3);
\texttt{bs\_cf\_sa} (3);
\texttt{bs\_cf\_ob\_sa} (2)
& 11 \\

Music &
\texttt{du\_ds} (3);
\texttt{am\_ds} (2);
\texttt{am\_du} (2);
\texttt{am\_ds\_du} (2)
& 9 \\

Health &
\texttt{cs\_cf} (6);
\texttt{cs\_tos} (6);
\texttt{tos\_cf} (6);
\texttt{cs\_cf\_tos} (6)
& 24 \\
\hline
Total & & 68 \\
\hline
\end{tabular}
\end{adjustbox}
\end{table}

\paragraph{Evaluation.}
We use the benchmark-defined task and deceptive-outcome evaluators over the recorded interaction trajectory. Task success is binary for each configuration. For a multi-pattern episode, the deceptive-outcome indicator is positive when at least one constituent dark-pattern condition succeeds, matching the definition of $D_i$ used in the main paper. The evaluator is applied after execution and is not available to the agent or defense at runtime.

\subsection{WebDecept}
\label{app:webdecept}

WebDecept contains 45 shopping tasks, each evaluated under seven deceptive scenarios, yielding $45\times7=315$ task--scenario configurations. Every task is evaluated under every scenario. Table~\ref{tab:webdecept-composition} summarizes the seven scenarios and the labels used in Figure~3.

\begin{table}[t]
\centering
\footnotesize
\setlength{\tabcolsep}{4pt}
\renewcommand{\arraystretch}{1.04}
\caption{WebDecept deceptive scenarios. Each scenario contains all 45 base shopping tasks.}
\label{tab:webdecept-composition}
\begin{adjustbox}{max width=\columnwidth}
\begin{tabular}{@{}lll r@{}}
\hline
Scenario & Label & Deceptive condition & $N$ \\
\hline
\texttt{popup}
& Popup
& Exit/checkout popup with deceptive continuation
& 45 \\

\texttt{banner}
& Banner
& Sticky deceptive discount advertisement
& 45 \\

\texttt{personalized\_popup}
& P-popup
& Task-conditioned deceptive popup
& 45 \\

\texttt{personalized\_banner}
& P-banner
& Task-conditioned deceptive banner
& 45 \\

\texttt{cart\_addons}
& Add-ons
& Additional donation item inserted into the cart
& 45 \\

\texttt{redirection}
& Redirect
& Fake continuation substitutes the intended path
& 45 \\

\texttt{price\_drift}
& Drift
& Cart price changes before commitment
& 45 \\
\hline
Total & & & 315 \\
\hline
\end{tabular}
\end{adjustbox}
\end{table}

We use the benchmark-defined task-success and deceptive-outcome evaluators for post-episode scoring. Task success determines whether the requested shopping outcome is completed, while the deceptive-outcome evaluator records whether the scenario-specific unsafe outcome occurs during the episode. These evaluators are not exposed to the base agent or Veer during execution.

\subsection{Feasibility-Aware WebDecept Subset}
\label{app:webdecept-feasible}

WebDecept contains two scenarios, \texttt{redirection} and \texttt{price\_drift}, whose benchmark ground truth does not specify a safe completion path once the deceptive condition is encountered. To separate this benchmark-level feasibility constraint from defense performance, we additionally evaluate the configurations from the remaining five scenarios, for which the benchmark ground truth specifies a safe completion path.

The resulting subset is

\[
\mathcal{C}_{\mathrm{feasible}}
=
\left\{
i \;\middle|\;
\mathrm{scenario}(i)
\notin
\{\texttt{redirection},\texttt{price\_drift}\}
\right\}.
\]

The criterion depends only on the benchmark scenario and is applied identically to every method. It yields $45\times5=225$ configurations.

\paragraph{Interpretation.}
The feasibility-aware subset is a secondary analysis and does not replace the
full 315-configuration WebDecept evaluation. We report the full benchmark
results for all methods and use the subset only to separate defense
performance from scenarios whose benchmark ground truth does not specify a
safe completion path after the deceptive condition is encountered. Subset
membership depends only on the benchmark scenario and is applied identically
to every method.
We therefore interpret the full-set TSR and the feasible-subset TSR together:
the former retains the benchmark's original feasibility constraints, while
the latter isolates configurations in which safe task completion is defined
by the benchmark ground truth.

\begin{table}[t]
\centering
\footnotesize
\setlength{\tabcolsep}{4pt}
\renewcommand{\arraystretch}{1.04}
\caption{Construction of the feasibility-aware WebDecept subset.}
\label{tab:webdecept-feasible}
\begin{tabular}{lrr}
\hline
Scenario & Full & Feasible \\
\hline
Popup    & 45 & 45 \\
Banner   & 45 & 45 \\
P-popup  & 45 & 45 \\
P-banner & 45 & 45 \\
Add-ons  & 45 & 45 \\
Redirect & 45 & 0 \\
Drift    & 45 & 0 \\
\hline
Total    & 315 & 225 \\
\hline
\end{tabular}
\end{table}

\subsection{Baseline Adaptations}
\label{app:baseline-adaptations}

We compare Veer with the unprotected base agent, deceptive-interface-specific defenses, and general agent-safety defenses. Each method is adapted only as needed to operate over the native observation and action interface of the corresponding benchmark.

\paragraph{No-Defense.}
No-Defense uses the native benchmark agent without defense-side screening or intervention.

\paragraph{ICP~\cite{cuvin2026dark}.}
ICP is instantiated as a static deceptive-interface warning inserted into the native actor context before action generation.

\paragraph{Guardrail~\cite{cuvin2026dark}.}
Guardrail performs deceptive-interface assessment over the current Web observation and supplies the resulting warning to the base agent.

\paragraph{DUDE-S2~\cite{zhang2026don}.}
We use the publicly reproducible Stage-2 defense, adapting its click-level review to the observation and click interfaces exposed by the two benchmark hosts.

\paragraph{Spotlighting~\cite{hines2024defending}.}
We apply Spotlighting by delimiting page content as untrusted input while preserving the native actor and interaction interface.

\paragraph{VIGIL~\cite{lin2026vigil}.}
We adapt VIGIL's sanitize--guide--audit workflow to the browser-action representation of each benchmark. Proposed actions that are not released by the defense are returned to the base agent for replanning.

\paragraph{SafePred~\cite{chen2026safepred}.}
We adapt SafePred's public runtime policy to the native observation and action interfaces. It predicts the consequence and risk of a proposed action and invokes base-agent replanning when the proposal does not satisfy the policy.

\paragraph{Adaptation scope.}
The adaptations are limited to mapping each defense to the observation,
prompt, and action interfaces exposed by the two benchmark hosts. They do not
add Veer's state-intervention mechanism to the baselines or change their
defense-side control point. The resulting implementations therefore preserve
the distinction between prompt-level warning, action review, consequence
screening, and state intervention that motivates the comparison.

\paragraph{Comparison protocol.}
Within each benchmark, all compared methods use the same task configurations, base agent, actor model, and base task-action allowance. Defense-specific auxiliary reasoning follows the corresponding method, while ordinary task execution remains subject to the common base-agent budget. No compared method is given benchmark task-success labels, dark-pattern labels, hidden evaluator outputs, or backend application state during runtime.
\section{Complete Results and Evaluation Protocol}
\label{app:complete-results}

This appendix reports the complete results underlying the analyses in the main paper. We provide the raw outcome counts for the overall comparisons, the exact per-condition DPSR values underlying Figure~3, and the complete agent--model results corresponding to Table~3.

\subsection{Evaluation Protocol}
\label{app:evaluation-protocol}

\paragraph{Evaluation units.}
TrickyArena-Single contains 88 task--condition configurations, TrickyArena-Multi contains 68 task--multi-condition configurations, and WebDecept contains 315 task--scenario configurations. The feasibility-aware WebDecept subset contains the 225 configurations from the five scenarios defined in Appendix~\ref{app:webdecept-feasible}. Within each comparison, all methods are evaluated on the same configuration identifiers.

\paragraph{Metrics.}
For configuration $i$, let $T_i$ denote binary task success and $D_i$ indicate whether at least one evaluated deceptive outcome occurs. We compute

\[
\mathrm{DPSR}
=
\frac{100}{N}\sum_{i=1}^{N}D_i,
\qquad
\mathrm{TSR}
=
\frac{100}{N}\sum_{i=1}^{N}T_i,
\]

and

\[
\mathrm{STC}
=
\frac{100}{N}\sum_{i=1}^{N}T_i(1-D_i).
\]

Importantly, STC is computed from the episode-level joint outcome
$T_i(1-D_i)$; it is not obtained by multiplying aggregate TSR by
$1-\mathrm{DPSR}$. We report DPSR and TSR separately so that safety and task
utility remain directly visible.

\paragraph{End-to-end evaluation.}
The reported metrics score the outcome of the complete runtime pipeline rather
than treating intermediate LLM judgments as independent evaluation samples.
Errors in consequence assessment or authorization therefore remain reflected
in the final episode outcome: an unsafe proposal that is incorrectly released
can increase DPSR, while an unnecessary intervention can reduce task success.
This evaluation directly measures the downstream effect of the assessment
procedure within the deployed defense.

For TrickyArena-Multi, $D_i$ is the logical OR over the constituent deceptive conditions active in configuration $i$. Thus, an episode contributes one positive instance to DPSR when at least one constituent dark-pattern outcome occurs.

The denominators are fixed at 88, 68, 315, and 225 for TrickyArena-Single, TrickyArena-Multi, full WebDecept, and the feasibility-aware WebDecept subset, respectively.

\subsection{Complete Overall Results}
\label{app:overall-results}

Tables~\ref{tab:single-complete}--\ref{tab:webdecept-feasible-complete} report the raw counts underlying Table~1 and Table~2 in the main paper. Here, \#D is the number of configurations with a deceptive outcome, \#T is the number with successful task completion, and \#STC is the number that complete the task without a deceptive outcome.

\begin{table}[t]
\centering
\footnotesize
\setlength{\tabcolsep}{4pt}
\renewcommand{\arraystretch}{1.04}
\caption{Complete TrickyArena-Single results ($N=88$).}
\label{tab:single-complete}
\begin{adjustbox}{max width=\columnwidth}
\begin{tabular}{@{}lrrrrrr@{}}
\hline
Method & \#D & \#T & \#STC & DPSR & TSR & STC \\
\hline
No-Defense   & 26 & 71 & 53 & 29.5 & 80.7 & 60.2 \\
ICP          & 15 & 71 & 61 & 17.0 & 80.7 & 69.3 \\
Guardrail    & 24 & 71 & 55 & 27.3 & 80.7 & 62.5 \\
DUDE-S2      & 30 & 76 & 50 & 34.1 & 86.4 & 56.8 \\
Spotlighting & 23 & 70 & 55 & 26.1 & 79.5 & 62.5 \\
VIGIL        & 14 & 63 & 52 & 15.9 & 71.6 & 59.1 \\
SafePred     & 25 & 76 & 56 & 28.4 & 86.4 & 63.6 \\
Veer         &  4 & 76 & 75 &  4.5 & 86.4 & 85.2 \\
\hline
\end{tabular}
\end{adjustbox}
\end{table}

\begin{table}[t]
\centering
\footnotesize
\setlength{\tabcolsep}{4pt}
\renewcommand{\arraystretch}{1.04}
\caption{Complete TrickyArena-Multi results ($N=68$).}
\label{tab:multi-complete}
\begin{adjustbox}{max width=\columnwidth}
\begin{tabular}{@{}lrrrrrr@{}}
\hline
Method & \#D & \#T & \#STC & DPSR & TSR & STC \\
\hline
No-Defense   & 37 & 45 & 26 & 54.4 & 66.2 & 38.2 \\
ICP          & 34 & 46 & 27 & 50.0 & 67.6 & 39.7 \\
Guardrail    & 32 & 48 & 29 & 47.1 & 70.6 & 42.6 \\
DUDE-S2      & 44 & 48 & 19 & 64.7 & 70.6 & 27.9 \\
Spotlighting & 40 & 41 & 18 & 58.8 & 60.3 & 26.5 \\
VIGIL        & 27 & 40 & 26 & 39.7 & 58.8 & 38.2 \\
SafePred     & 38 & 54 & 25 & 55.9 & 79.4 & 36.8 \\
Veer         & 10 & 49 & 46 & 14.7 & 72.1 & 67.6 \\
\hline
\end{tabular}
\end{adjustbox}
\end{table}

\begin{table}[t]
\centering
\footnotesize
\setlength{\tabcolsep}{4pt}
\renewcommand{\arraystretch}{1.04}
\caption{Complete WebDecept results over all seven scenarios ($N=315$).}
\label{tab:webdecept-complete}
\begin{adjustbox}{max width=\columnwidth}
\begin{tabular}{@{}lrrrrrr@{}}
\hline
Method & \#D & \#T & \#STC & DPSR & TSR & STC \\
\hline
No-Defense   & 118 & 150 &  90 & 37.5 & 47.6 & 28.6 \\
ICP          &  84 & 138 & 100 & 26.7 & 43.8 & 31.7 \\
Guardrail    &  39 & 111 &  93 & 12.4 & 35.2 & 29.5 \\
DUDE-S2      & 113 & 175 &  99 & 35.9 & 55.6 & 31.4 \\
Spotlighting & 112 & 141 &  86 & 35.6 & 44.8 & 27.3 \\
VIGIL        &  30 &  32 &  17 &  9.5 & 10.2 &  5.4 \\
SafePred     & 120 & 183 & 122 & 38.1 & 58.1 & 38.7 \\
Veer         &   1 & 129 & 129 &  0.3 & 41.0 & 41.0 \\
\hline
\end{tabular}
\end{adjustbox}
\end{table}

\begin{table}[t]
\centering
\footnotesize
\setlength{\tabcolsep}{4pt}
\renewcommand{\arraystretch}{1.04}
\caption{Complete results on the feasibility-aware WebDecept subset ($N=225$).}
\label{tab:webdecept-feasible-complete}
\begin{adjustbox}{max width=\columnwidth}
\begin{tabular}{@{}lrrrrrr@{}}
\hline
Method & \#D & \#T & \#STC & DPSR & TSR & STC \\
\hline
No-Defense   & 40 & 113 &  90 & 17.8 & 50.2 & 40.0 \\
ICP          & 13 & 107 & 100 &  5.8 & 47.6 & 44.4 \\
Guardrail    & 16 & 102 &  93 &  7.1 & 45.3 & 41.3 \\
DUDE-S2      & 38 & 122 &  99 & 16.9 & 54.2 & 44.0 \\
Spotlighting & 39 & 107 &  86 & 17.3 & 47.6 & 38.2 \\
VIGIL        &  7 &  20 &  17 &  3.1 &  8.9 &  7.6 \\
SafePred     & 44 & 145 & 122 & 19.6 & 64.4 & 54.2 \\
Veer         &  0 & 129 & 129 &  0.0 & 57.3 & 57.3 \\
\hline
\end{tabular}
\end{adjustbox}
\end{table}

\subsection{Per-Condition Results}
\label{app:per-condition}

Tables~\ref{tab:single-condition-dpsr} and~\ref{tab:webdecept-scenario-dpsr} provide the exact DPSR values underlying Figure~3. Each cell reports DPSR in percent followed by the raw number of deceptive outcomes in parentheses.

\begin{table*}[t]
\centering
\scriptsize
\setlength{\tabcolsep}{3pt}
\renewcommand{\arraystretch}{1.03}
\caption{TrickyArena-Single DPSR by condition. Each cell reports DPSR (\%) with raw \#D in parentheses.}
\label{tab:single-condition-dpsr}
\begin{adjustbox}{max width=\textwidth}
\begin{tabular}{@{}lrrrrrrrrr@{}}
\hline
Condition & $N$ & No-Def. & ICP & Guard. & DUDE & Spot. & VIGIL & SafePred & Veer \\
\hline
shop/p1  & 4 & 0.0(0)   & 0.0(0)   & 0.0(0)   & 0.0(0)   & 0.0(0)   & 0.0(0)   & 0.0(0)   & 0.0(0) \\
shop/t1  & 4 & 0.0(0)   & 0.0(0)   & 0.0(0)   & 0.0(0)   & 0.0(0)   & 0.0(0)   & 0.0(0)   & 0.0(0) \\
shop/t2  & 4 & 75.0(3)  & 0.0(0)   & 25.0(1)  & 75.0(3)  & 25.0(1)  & 0.0(0)   & 100.0(4) & 25.0(1) \\
shop/t3  & 4 & 0.0(0)   & 0.0(0)   & 0.0(0)   & 25.0(1)  & 0.0(0)   & 0.0(0)   & 0.0(0)   & 0.0(0) \\
shop/t4  & 4 & 0.0(0)   & 0.0(0)   & 0.0(0)   & 0.0(0)   & 0.0(0)   & 0.0(0)   & 0.0(0)   & 0.0(0) \\
shop/t5  & 4 & 0.0(0)   & 0.0(0)   & 0.0(0)   & 25.0(1)  & 0.0(0)   & 0.0(0)   & 0.0(0)   & 0.0(0) \\
shop/t6  & 4 & 0.0(0)   & 0.0(0)   & 0.0(0)   & 25.0(1)  & 0.0(0)   & 0.0(0)   & 0.0(0)   & 0.0(0) \\
shop/t7  & 4 & 0.0(0)   & 0.0(0)   & 0.0(0)   & 0.0(0)   & 0.0(0)   & 0.0(0)   & 0.0(0)   & 0.0(0) \\
shop/t8  & 4 & 0.0(0)   & 0.0(0)   & 0.0(0)   & 0.0(0)   & 0.0(0)   & 0.0(0)   & 0.0(0)   & 0.0(0) \\
shop/p2  & 4 & 100.0(4) & 50.0(2)  & 100.0(4) & 100.0(4) & 100.0(4) & 75.0(3)  & 100.0(4) & 0.0(0) \\
shop/s   & 5 & 0.0(0)   & 0.0(0)   & 0.0(0)   & 0.0(0)   & 0.0(0)   & 0.0(0)   & 0.0(0)   & 0.0(0) \\
shop/w   & 5 & 20.0(1)  & 0.0(0)   & 20.0(1)  & 20.0(1)  & 0.0(0)   & 0.0(0)   & 0.0(0)   & 0.0(0) \\
news/bs  & 3 & 0.0(0)   & 0.0(0)   & 0.0(0)   & 0.0(0)   & 0.0(0)   & 0.0(0)   & 0.0(0)   & 0.0(0) \\
news/ob  & 3 & 0.0(0)   & 0.0(0)   & 0.0(0)   & 0.0(0)   & 0.0(0)   & 0.0(0)   & 0.0(0)   & 0.0(0) \\
news/sa  & 3 & 0.0(0)   & 0.0(0)   & 0.0(0)   & 0.0(0)   & 0.0(0)   & 0.0(0)   & 0.0(0)   & 0.0(0) \\
news/cf  & 3 & 100.0(3) & 100.0(3) & 100.0(3) & 100.0(3) & 100.0(3) & 100.0(3) & 100.0(3) & 100.0(3) \\
music/am & 2 & 0.0(0)   & 0.0(0)   & 0.0(0)   & 0.0(0)   & 0.0(0)   & 0.0(0)   & 0.0(0)   & 0.0(0) \\
music/ds & 3 & 0.0(0)   & 100.0(3) & 0.0(0)   & 33.3(1)  & 0.0(0)   & 0.0(0)   & 0.0(0)   & 0.0(0) \\
music/du & 3 & 100.0(3) & 0.0(0)   & 100.0(3) & 100.0(3) & 100.0(3) & 66.7(2)  & 66.7(2)  & 0.0(0) \\
health/cs  & 6 & 100.0(6) & 100.0(6) & 100.0(6) & 100.0(6) & 100.0(6) & 100.0(6) & 100.0(6) & 0.0(0) \\
health/tos & 6 & 100.0(6) & 16.7(1)  & 100.0(6) & 100.0(6) & 100.0(6) & 0.0(0)   & 100.0(6) & 0.0(0) \\
health/cf  & 6 & 0.0(0)   & 0.0(0)   & 0.0(0)   & 0.0(0)   & 0.0(0)   & 0.0(0)   & 0.0(0)   & 0.0(0) \\
\hline
\end{tabular}
\end{adjustbox}
\end{table*}

On TrickyArena-Single, Veer records zero deceptive outcomes in 20 of the 22 conditions and achieves the lowest or tied-lowest DPSR in 21 conditions. Its unweighted mean per-condition DPSR is 5.7\%, compared with its configuration-weighted overall DPSR of 4.5\%.

\begin{table*}[t]
\centering
\scriptsize
\setlength{\tabcolsep}{3.5pt}
\renewcommand{\arraystretch}{1.03}
\caption{WebDecept DPSR by deceptive scenario. Each cell reports DPSR (\%) with raw \#D in parentheses.}
\label{tab:webdecept-scenario-dpsr}
\begin{adjustbox}{max width=\textwidth}
\begin{tabular}{@{}lrrrrrrrrr@{}}
\hline
Scenario & $N$ & No-Def. & ICP & Guard. & DUDE & Spot. & VIGIL & SafePred & Veer \\
\hline
Popup    & 45 & 4.4(2)  & 4.4(2)  & 6.7(3)  & 0.0(0)  & 6.7(3)  & 2.2(1)  & 6.7(3)  & 0.0(0) \\
Banner   & 45 & 0.0(0)  & 0.0(0)  & 0.0(0)  & 0.0(0)  & 0.0(0)  & 0.0(0)  & 0.0(0)  & 0.0(0) \\
P-popup  & 45 & 0.0(0) & 0.0(0) & 0.0(0) & 0.0(0)
         & 0.0(0) & 2.2(1) & 2.2(1) & 0.0(0) \\
P-banner & 45 & 2.2(1)  & 2.2(1)  & 0.0(0)  & 2.2(1)  & 2.2(1)  & 0.0(0)  & 2.2(1)  & 0.0(0) \\
Add-ons  & 45 & 82.2(37)& 22.2(10)& 28.9(13)& 82.2(37)& 77.8(35)& 11.1(5)& 86.7(39)& 0.0(0) \\
Redirect & 45 & 86.7(39)& 80.0(36)& 48.9(22)& 82.2(37)& 82.2(37)& 26.7(12)& 84.4(38)& 2.2(1) \\
Drift    & 45 & 86.7(39)& 77.8(35)& 2.2(1)  & 84.4(38)& 80.0(36)& 24.4(11)& 84.4(38)& 0.0(0) \\
\hline
\end{tabular}
\end{adjustbox}
\end{table*}

Across WebDecept, Veer records one deceptive outcome among 315 configurations. Its unweighted mean per-scenario DPSR is 0.3\%, and its maximum scenario DPSR is 2.2\%.

\subsection{Agent--Model Results}
\label{app:agent-model-results}

Table~\ref{tab:agent-model-complete} reports the complete system-configuration results corresponding to Table~3 in the main paper. Each cell gives Veer's STC followed by the absolute improvement over the matched No-Defense configuration in percentage points.

\begin{table*}[t]
\centering
\footnotesize
\setlength{\tabcolsep}{6pt}
\renewcommand{\arraystretch}{1.05}
\caption{Complete STC results across agent--model configurations. Parentheses denote absolute improvement over the matched No-Defense configuration in percentage points.}
\label{tab:agent-model-complete}
\begin{adjustbox}{max width=\textwidth}
\begin{tabular}{@{}llccc@{}}
\hline
Agent & Model & TrickyArena-Single & TrickyArena-Multi & WebDecept \\
\hline
Default & GPT-5.6-Luna
& 85.2 (+25.0) & 67.6 (+29.4) & 41.0 (+12.4) \\

Default & DeepSeek-V4-Flash
& 68.2 (+12.5) & 44.1 (+10.3) & 40.0 (+4.8) \\

Codex & GPT-5.6-Luna
& 75.0 (+28.4) & 54.4 (+25.0) & 45.4 (+5.1) \\

Codex & DeepSeek-V4-Flash
& 85.2 (+20.4) & 55.9 (+14.7) & 53.7 (+15.9) \\
\hline
\end{tabular}
\end{adjustbox}
\end{table*}

Veer improves STC over the matched No-Defense setting in all 12 agent--model--benchmark comparisons, consistent with the robustness result reported in the main paper.
\section{Additional Ablation and Trajectory Analysis}
\label{app:additional-ablations}

This appendix reports the complete RQ3 ablation results, an additional one-step-intervention ablation, and representative trajectory analyses. All ablations use TrickyArena-Single with $N=88$ configurations per variant. Unless explicitly removed by an ablation, the variants retain the same user
tasks, base agent, actor model, task-action allowance, and surrounding Veer
runtime components.

\subsection{Complete Ablation Results}
\label{app:complete-ablation}

Table~\ref{tab:complete-ablation} decomposes each episode into four mutually exclusive outcomes: safe task completion ($T=1,D=0$), task completion with a dark-pattern outcome ($T=1,D=1$), task failure without a dark-pattern outcome ($T=0,D=0$), and task failure with a dark-pattern outcome ($T=0,D=1$).

\begin{table*}[t]
\centering
\footnotesize
\setlength{\tabcolsep}{4pt}
\renewcommand{\arraystretch}{1.05}
\caption{Complete RQ3 ablation results on TrickyArena-Single ($N=88$ per variant). $\Delta$STC is measured relative to Full Veer.}
\label{tab:complete-ablation}
\begin{adjustbox}{max width=\textwidth}
\begin{tabular}{@{}lrrrrrrrr@{}}
\hline
Variant
& Safe
& T+DP
& Fail, no-DP
& Fail+DP
& DPSR (\%)
& TSR (\%)
& STC (\%)
& $\Delta$STC (pp) \\
\hline
Full Veer
& 75 & 1 & 9 & 3
& 4.5 & 86.4 & 85.2 & -- \\

w/o State Intervention
& 44 & 1 & 34 & 9
& 11.4 & 51.1 & 50.0 & $-35.2$ \\

Reactive Intervention
& 71 & 4 & 11 & 2
& 6.8 & 85.2 & 80.7 & $-4.5$ \\

w/o Temporal Evidence
& 68 & 3 & 15 & 2
& 5.7 & 80.7 & 77.3 & $-8.0$ \\
\hline
\end{tabular}
\end{adjustbox}
\end{table*}

\paragraph{State intervention.}
The \emph{w/o State Intervention} variant replaces corrective state intervention with blocking. When Veer identifies an unauthorized consequence, the proposed action is rejected without issuing a corrective browser action. This substantially reduces task progress: TSR decreases from 86.4\% to 51.1\%, and STC decreases by 35.2 percentage points.

\paragraph{Prospective rollout.}
The \emph{Reactive Intervention} variant retains the same intervention objective, temporal evidence, preservation constraints, grounding, and runtime verification as Full Veer. It removes complete prospective rollout: Veer selects one currently grounded corrective transition, executes it, observes the resulting state, and determines the next transition only if further correction is required. TSR remains close to Full Veer, while DPSR increases from 4.5\% to 6.8\% and STC decreases from 85.2\% to 80.7\%.

\paragraph{Temporal evidence.}
The \emph{w/o Temporal Evidence} variant removes the retained temporal evidence used by Veer across protected interaction steps while preserving the current Web observation and the base agent's normal interaction context. Removing this evidence reduces TSR from 86.4\% to 80.7\% and STC from 85.2\% to 77.3\%, showing the value of retaining consequence-relevant state across multi-step interaction.

\subsection{Additional One-Step Ablation}
\label{app:one-step-ablation}

The main Reactive Intervention ablation removes complete prospective rollout while still allowing successive corrective transitions to be constructed after observing each intermediate state. We further evaluate a stricter \emph{One-Step Intervention} variant that permits only one local corrective browser action per interception and then returns control to the base-agent loop.

\begin{table}[t]
\centering
\footnotesize
\setlength{\tabcolsep}{4pt}
\renewcommand{\arraystretch}{1.05}
\caption{Additional one-step-intervention ablation on TrickyArena-Single ($N=88$).}
\label{tab:one-step-ablation}
\begin{adjustbox}{max width=\columnwidth}
\begin{tabular}{@{}lrrrr@{}}
\hline
Variant & DPSR (\%) & TSR (\%) & STC (\%) & $\Delta$STC (pp) \\
\hline
Full Veer            & 4.5  & 86.4 & 85.2 & -- \\
One-Step Intervention& 14.8 & 68.2 & 61.4 & $-23.9$ \\
\hline
\end{tabular}
\end{adjustbox}
\end{table}

Unlike Full Veer, this variant does not represent or verify a complete multi-step correction with cross-step dependencies. Its STC decreases from 85.2\% to 61.4\%, while DPSR increases from 4.5\% to 14.8\%. Together with the Reactive Intervention result, this shows that preserving the structure of a multi-step correction is important when safe intervention requires dependent state transitions.

\subsection{Intervention-Trajectory Analysis}
\label{app:trajectory-analysis}

We next examine the prospective trajectories admitted by Full Veer to characterize the structure of state intervention in practice.

\begin{table}[t]
\centering
\footnotesize
\setlength{\tabcolsep}{6pt}
\renewcommand{\arraystretch}{1.05}
\caption{Length of admitted prospective intervention trajectories.}
\label{tab:trajectory-length}
\begin{tabular}{lrr}
\hline
Planned length $K$ & Episodes & Trajectories \\
\hline
1 & 15 & 17 \\
4 & 6  & 7  \\
\hline
Total & 21 & 24 \\
\hline
\end{tabular}
\end{table}

Among episodes in which Veer admits an intervention, 15 use a one-transition trajectory and six use a four-transition trajectory. Across the 24 admitted trajectories, the mean planned length is 1.88 transitions, the median is one, and the maximum is four. The presence of four-transition trajectories confirms that the evaluated interventions include dependent multi-step corrections rather than only isolated local actions.

The executed corrective actions span four planner-level operation types.

\begin{table}[t]
\centering
\footnotesize
\setlength{\tabcolsep}{8pt}
\renewcommand{\arraystretch}{1.05}
\caption{Executed corrective operations in Full Veer.}
\label{tab:correction-operations}
\begin{tabular}{lr}
\hline
Corrective operation & Executed actions \\
\hline
\texttt{turn\_off} & 20 \\
\texttt{decline}   & 12 \\
\texttt{save}      & 6 \\
\texttt{click}     & 4 \\
\hline
Total               & 42 \\
\hline
\end{tabular}
\end{table}

These operations cover both direct state reduction, such as disabling or declining an unwanted state, and transitions that commit an already constructed correction, such as saving the resulting settings.

\subsection{Representative Intervention Trajectories}
\label{app:representative-trajectories}

We provide two representative episodes that illustrate state correction with task preservation and guarded verification with bounded replanning.

\paragraph{Case 1: consent correction with task preservation.}
In configuration \texttt{health\_tos\_80}, the user asks the agent to retrieve the date of the last flu shot and write it to the scratchpad. The base agent proposes an action that would commit to broad health-data consent unrelated to the requested task. Veer constructs an intervention objective that removes the pending consent while preserving the task-relevant scratchpad capability.

The prospective trajectory contains a single corrective transition:

\begin{table}[t]
\centering
\footnotesize
\setlength{\tabcolsep}{6pt}
\renewcommand{\arraystretch}{1.05}
\caption{Representative consent correction for \texttt{health\_tos\_80}.}
\label{tab:health-tos-trajectory}
\begin{tabular}{llll}
\hline
Step & Operation & Expected state & Dependency \\
\hline
$\tau_1$ & decline consent & declined & -- \\
\hline
\end{tabular}
\end{table}

After the correction, Veer verifies that the consent state has been removed while the task-relevant capability remains available. Control returns to the base agent, which retrieves the requested medical-record information and writes it to the scratchpad. The episode finishes with $T=1,D=0$.

This case illustrates the role of $P_t$: state correction removes the unauthorized consequence while retaining state needed for the original task.

\paragraph{Case 2: terminal verification and bounded replanning.}
In configuration \texttt{shop\_p2\_38}, the user asks for the description of a laptop. The base agent proposes accepting cookie consent that is unnecessary for the requested information-retrieval task. Veer constructs an objective that prevents committing this consent while preserving product retrieval and scratchpad-entry capability.

The initial prospective correction opens the cookie-options interface. The browser action succeeds locally, but the resulting state does not yet satisfy the intervention objective. Veer therefore retains the same objective and constructs a new corrective continuation:

\begin{table}[t]
\centering
\footnotesize
\setlength{\tabcolsep}{6pt}
\renewcommand{\arraystretch}{1.05}
\caption{Bounded intervention replan for \texttt{shop\_p2\_38}.}
\label{tab:cookie-replan}
\begin{tabular}{lll}
\hline
Attempt & Corrective transition & Result \\
\hline
Initial & Open cookie options & Objective unresolved \\
Replan  & Decline cookie consent & Verified \\
\hline
\end{tabular}
\end{table}

After the corrected state is verified, control returns to the base agent, which resumes product retrieval and writes the requested description to the scratchpad. The episode finishes with $T=1,D=0$.

This case illustrates why Veer verifies the resulting state against the intervention objective after browser execution: a locally successful interaction does not by itself establish that the intended state correction has been completed.
\section{Scope and Limitations}
\label{app:limitations}

Veer is designed as an agent-side runtime defense for Web-agent execution under deceptive interfaces. Its protection operates over the same black-box browser interface available to the base agent: it reasons from observable Web evidence, intervenes through grounded browser interactions, and derives authorization from the original user task. This section summarizes the resulting applicability boundary.

\paragraph{Observable-state dependence.}
Veer constructs task-relevant Web state from the current observation and retained temporal evidence. Consequently, an intervention must be grounded in state that has an observable manifestation in the Web interaction. Hidden application state that cannot be inferred from available Web evidence does not directly participate in consequence assessment or intervention planning.
Accordingly, Veer targets consequences whose relevant pre-commit state has an
observable Web manifestation; effects determined entirely by hidden
server-side state with no observable evidence fall outside the current
intervention model.

\paragraph{Browser-reachable correction.}
State intervention requires a browser-level path from the current state toward the target state. Dynamic interfaces, unavailable controls, or unexpected transition effects can invalidate a prospective trajectory. Veer addresses these cases through live grounding, dependency checks, and post-transition verification, and replans under the same intervention objective when a valid continuation remains available.

\paragraph{Task-grounded authorization.}
Veer derives authorization from the original user instruction and does not expand it using Web content encountered during execution. Runtime observations can ground task references to concrete objects and facts, while the authorization boundary remains fixed. Tasks whose intent is insufficiently specified can therefore leave some proposed consequences uncertain.

\paragraph{Task-state preservation.}
The preservation component of the intervention objective covers task-relevant state and capabilities that can be identified from the interaction evidence available to Veer. Guarded execution verifies these declared preservation requirements together with the target corrected state before returning control to the base agent.

\paragraph{Base-agent dependence.}
Veer intervenes on consequence-causing Web state and then returns control to the base agent. It does not replace the base agent's ordinary task planner. Navigation, information retrieval, and completion of the remaining user task therefore continue to depend on the underlying agent after a successful intervention.

\paragraph{Evaluation scope.}
Our evaluation covers TrickyArena-Single, TrickyArena-Multi, and WebDecept, spanning multiple deceptive-interface mechanisms, application domains, agent implementations, and model configurations. These experiments establish the effectiveness of consequence-guided state intervention in the evaluated settings; extending the evaluation to additional Web environments and interaction settings remains future work.

\end{document}